\documentclass[journal]{IEEEtran}

\usepackage[numbers,sort&compress]{natbib} % 多个cite合并，简写多个数字
\usepackage{float}

\usepackage{multirow}
\usepackage{threeparttable}
\usepackage{booktabs}

\usepackage[ruled,linesnumbered]{algorithm2e} 
\usepackage{amsmath}
\usepackage{amsfonts}
\usepackage{amssymb}  

\usepackage{graphicx}
\graphicspath{{images/}}                     %表示在当前目录下存放有一个图片，下次插入图片从此路径开始

\usepackage{hyperref}

\usepackage[utf8]{inputenc}

\usepackage{soul}
\soulregister\cite7
\soulregister\citep7
\soulregister\citet7
\soulregister\ref7
\soulregister\pageref7
\usepackage{color, xcolor} % 颜色包，color 必须导入，xcolor 建议导入

\begin{document}
%

% ******************************paper title****************************** 
\title{Continuous Cross-resolution Remote Sensing Image Change Detection}

% ******************************author names****************************** 

\author{
Hao~Chen,
        Haotian~Zhang,
        Keyan~Chen,
        Chenyao~Zhou,
        % ,
        Song~Chen,
        Zhengxia~Zou
        and 
        Zhenwei~Shi$^{*}$,~\IEEEmembership{Member,~IEEE}
        % <-this % stops a space
% ******************************IEEE memberships****************************** 
\thanks{The work was supported by the National Key Research and Development Program of China (Grant No. 2022ZD0160401), the National Natural Science Foundation of China under Grant 62125102, the Beijing Natural Science Foundation under Grant JL23005, and the Fundamental Research Funds for the Central Universities. \emph{(Corresponding Author: Zhenwei Shi (shizhenwei@buaa.edu.cn))}.}

\thanks{Hao Chen, Haotian Zhang, Keyan Chen, Chenyao Zhou, and Zhenwei Shi are with the Image Processing Center, School of Astronautics, Beihang University, Beijing 100191, China, and with the Beijing Key Laboratory of Digital Media, Beihang University, Beijing 100191, China, and with the State Key Laboratory of Virtual Reality Technology and Systems, Beihang University, Beijing 100191, China, and also with the Shanghai Artificial Intelligence Laboratory, Shanghai 200232, China.}

\thanks{Song Chen is with the Department of Journalism and Communications, Jeonuk National University, Jeonju-si 54896, South Korea.}

\thanks{Zhengxia Zou is with the Department of Guidance, Navigation and Control, School of Astronautics, Beihang University, Beijing 100191, China, and also with the Shanghai Artificial Intelligence Laboratory, Shanghai 200232, China.}
}

% }

% ******************************The paper headers****************************** 

% \markboth{Journal of \LaTeX\ Class Files,~Vol.~14, No.~8, August~2015}%
% {Shell \MakeLowercase{\textit{et al.}}: Bare Demo of IEEEtran.cls for IEEE Journals}

\maketitle

% ******************************abstract******************************
\begin{abstract}
Most contemporary supervised Remote Sensing (RS) image Change Detection (CD) approaches are customized for equal-resolution bitemporal images. Real-world applications raise the need for cross-resolution change detection, aka, CD based on bitemporal images with different spatial resolutions. Given training samples of a fixed bitemporal resolution difference (ratio) between the high-resolution (HR) image and the low-resolution (LR) one, current cross-resolution methods may fit a certain ratio but lack adaptation to other resolution differences. Toward continuous cross-resolution CD, we propose scale-invariant learning to enforce the model consistently predicting HR results given synthesized samples of varying resolution differences. Concretely, we synthesize blurred versions of the HR image by random downsampled reconstructions to reduce the gap between HR and LR images. We introduce coordinate-based representations to decode per-pixel predictions by feeding the coordinate query and corresponding multi-level embedding features into an MLP that implicitly learns the shape of land cover changes, therefore benefiting recognizing blurred objects in the LR image. Moreover, considering that spatial resolution mainly affects the local textures, we apply local-window self-attention to align bitemporal features during the early stages of the encoder. Extensive experiments on two synthesized and one real-world different-resolution CD datasets verify the effectiveness of the proposed method. Our method significantly outperforms several vanilla CD methods and two cross-resolution CD methods on the three datasets both in in-distribution and out-of-distribution settings. The empirical results suggest that our method could yield relatively consistent HR change predictions regardless of varying bitemporal resolution ratios. Our code is available at \url{https://github.com/justchenhao/SILI_CD}.

\end{abstract}

% ****************************** keywords****************************** 
\begin{IEEEkeywords}
High-resolution remote sensing image,
Cross-resolution change detection,
Scale-invariant learning,
Implicit neural representation,
Attention mechanism.
\end{IEEEkeywords}

\IEEEpeerreviewmaketitle

%  ******************************Introduction******************************
\section{Introduction}
\label{sec:intro}

\IEEEPARstart{R}{emote} sensing (RS) image change detection (CD) refers to identifying changes of interest objects or phenomena in the scene by comparing co-registered multi-temporal RS images taken at different times in the same geographical area \cite{SINGH1989}. Change detection could be applied to various applications, e.g., urban planning \cite{Chen2020e}, disaster management \cite{Xu2019}, agricultural surveys \cite{Bruzzone2000}, and environmental monitoring \cite{Bem2020}.

The availability of high-resolution (HR) remote sensing (RS) images enables monitoring changes on Earth’s surface at a fine scale. Existing deep learning-based techniques, such as convolutional neural networks (CNNs) \cite{He2016} and vision transformers \cite{Dosovitskiy2020}, are widely applied in RS CD \cite{Shi2020}. Despite promising results, most existing supervised CD approaches are customized for handling equal-resolution bitemporal images and are insufficient to adapt to cross-resolution conditions, where bitemporal images have different resolutions.

Real-world applications raise the need for change recognition based on multi-temporal images across resolutions. We identify roughly two scenarios: 
1) the long-term CD task, with a relatively low-resolution (LR) pre-event image and an HR post-event one considering earlier satellite observations (e.g., decades before) have relatively lower spatial resolution than those obtained by current satellite sensors; 2) the event/disaster rapid response task, with an archived HR pre-event image of a certain area and a relatively LR post-event image, considering the lack of real-time availability of HR satellite data, due to its smaller spatial coverage and longer revisit period, compared to LR data.

\begin{figure}
        \centering
        \includegraphics[width=0.5\textwidth]{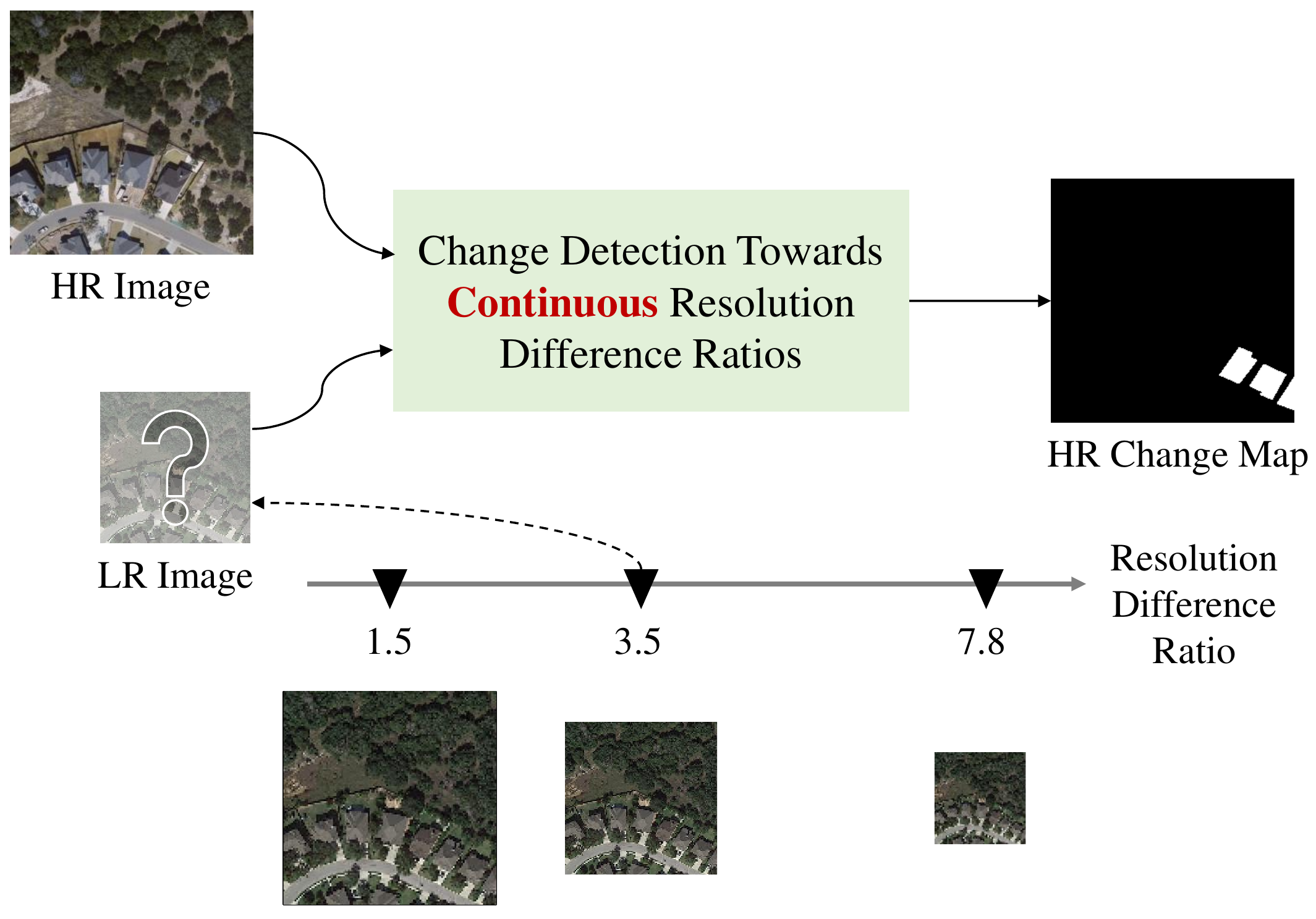}   
        \caption{Illustration of continuous cross-resolution change detection, i.e., CD towards varying resolution difference ratios between the HR image and the relatively LR image.
        }
        \label{fig:continous_cd}
\end{figure}

To handle the cross-resolution RS CD, aka., change detection based on bitemporal images with different spatial resolutions, most current methods \cite{Tian2022a, Liu2022g, Wang2022j, Tu2017, Zhang2016, Liu2022f, Shao2021} align the two inputs in the image space, either by downsampling the HR image \cite{Tu2017, Zhang2016} or upsampling the LR image in a fixed (e.g., bilinear/cubic interpolation) \cite{Wang2022j} or a learnable manner \cite{Liu2022g, Tian2022a}. Recent attempts \cite{Liu2022f, Shao2021, Zheng2022a} align the bitemporal resolution differences in the feature space, e.g., upsampling the feature map of the LR image by considering that of the HR one \cite{Liu2022f}.

Despite current progress in cross-resolution CD, a model \cite{Tian2022a, Liu2022g, Wang2022j, Tu2017, Zhang2016, Liu2022f, Shao2021} trained with a fixed resolution difference (i.e., difference ratio, e.g., 4 or 8) may work well for a certain condition, but may not be suitable for situations of other resolution differences, which limits its real-world applications. To fill this gap, different from existing approaches that are specifically designed for a fixed bitemporal resolution difference, we explore a continuous cross-resolution CD method that enables a single model to adapt arbitrary difference ratios between bitemporal images. Different from the traditional cross-resolution CD task that applies a fixed resolution difference for assessment, the continuous cross-resolution CD task evaluates the CD model on validation/testing samples with varying bitemporal resolution difference ratios that may be different from that of the training samples.
As shown in Fig. \ref{fig:continous_cd}, given an HR image and a relatively LR image, our goal is to obtain the HR change map regardless of resolution difference ratios.

To achieve this, we propose a scale-invariant training pipeline to learn an embedding space that is insensitive to scale changes of input images. Given original training samples with a fixed resolution ratio, we synthesize samples with random resolution differences by downsampling HR images and swapping bitemporal regions. We enforce the model yielding HR predictions for these synthesized samples, thus improving the ability to adapt different resolution ratios. We then incorporate the coordinate-based method, namely implicit neural representation (INR) \cite{Mildenhall2022}, to decode pixel-wise change information from the embedding space and corresponding pixel positions. 
Specifically, a multi-level feature embedding space is learned for a trade-off between semantic accuracy and spatial details \cite{Long2015}. Different from existing CD methods that employ sophisticated multi-level feature fusion (e.g., UNet \cite{Peng2019, Bem2020, Daudt2018, Hou2020, Zhang2020b, Liu2019b, Papadomanolaki2019, Pei2022, Li2022c, Chen2021a} or FPN\cite{Zhang2020a, Bao2020, Jiang2020}) to yield HR predictions, our coordinate-based approach implicitly learns land-cover shapes that may benefit handling LR images with blurry low-quality objects.
% We incorporate the scale information (i.e., pixel area with respect to HR image) of each level into the decoder network to align the multi-level features.
Furthermore, we propose bitemporal interaction on the early-level features to further fill the resolution gap by applying transformers \cite{Vaswani2017} to model correlations between bitemporal pixels within the local windows on the feature maps. Motivated by the fact that spatial resolution differences directly affect the local textures and image details are locally correlated without long-range dependency \cite{Guo2022}, only local information of the LR and HR patches may be sufficient to model correlations between cross-resolution pixels.

\begin{figure*}
        \centering
        \includegraphics[width=1\textwidth]{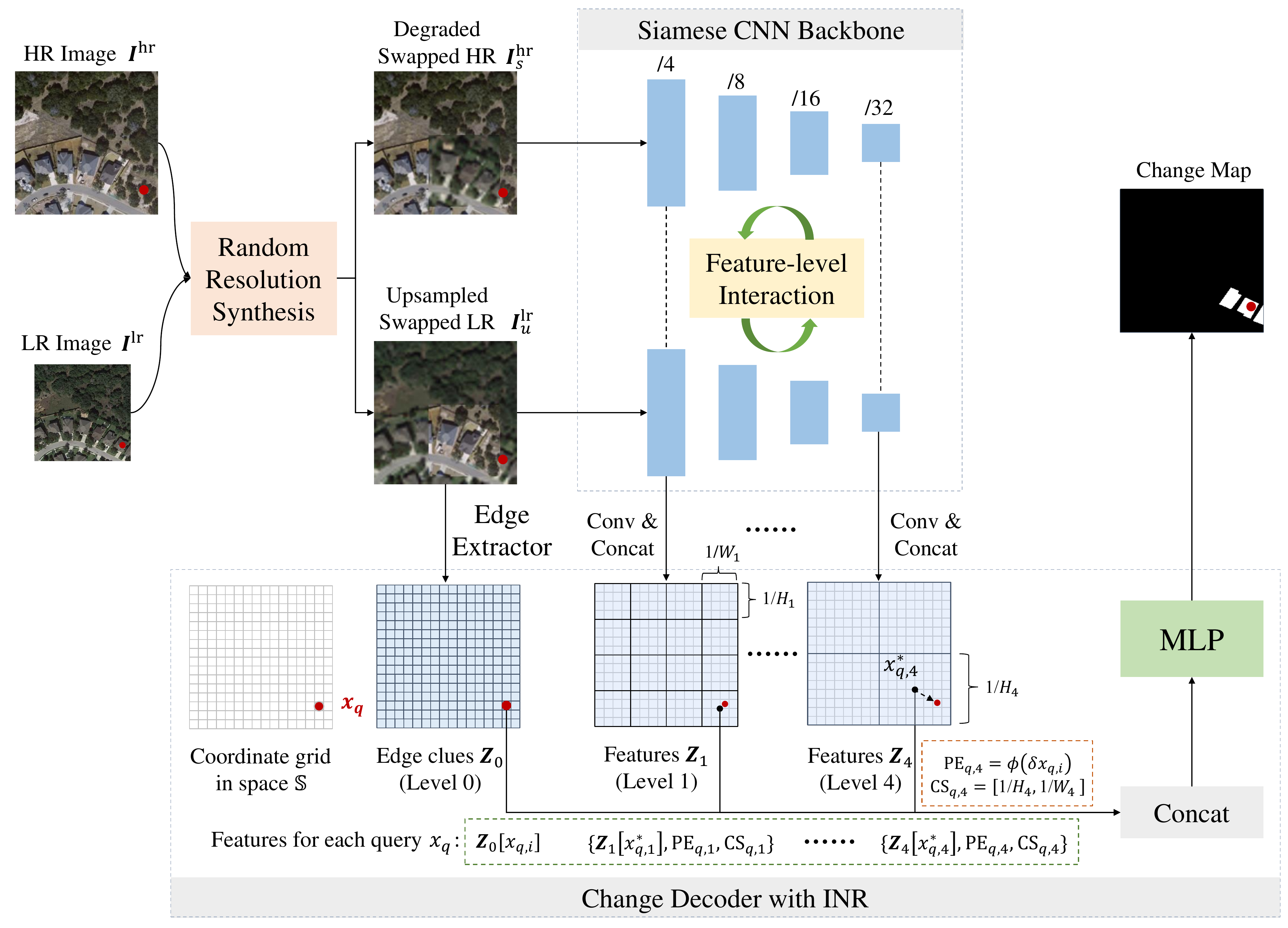}   
        \caption{Illustration of the overall pipeline of the proposed Scale-Invariant Learning with Implicit neural networks (SILI). We aim to learn a multi-level embedding space invariant to resolution differences across bitemporal images by enforcing the model generating HR change masks for synthesized samples with random resolution difference ratios. We leverage implicit neural representations that encapsulate the local mask shape to decode the HR mask from dense coordinate queries and corresponding multi-level features, including learnable edge clues. Note that we calculate positional encoding (PE) of the relative coordinate between the query and corresponding feature cell. The cell scale (CS) of each level is also fed into the decoder.
        }
        \label{fig:overall}
\end{figure*}

The contribution of our work can be summarised as follows:
\begin{itemize}
    \item We propose a scale-invariant methodology whereby an embedding space insensitive to scale changes is learned for cross-resolution RS image CD. Unlike extant approaches that are tailored to specific difference ratios between bitemporal resolutions, our method is capable of adapting to continuous resolution difference ratios. 
    \item We introduce coordinate-based representations to decode the HR change mask from the embedding space by implicitly learning the shape of objects of interest, therefore benefiting recognizing blurred objects in the LR image. Moreover, we incorporate local in-window interactions between bitemporal features to equip the model to better adapt to resolution disparities across bitemporal images.
    \item Extensive experiments on two synthetic and one real-world cross-resolution CD datasets validate the effectiveness of the proposed method. Our approach outperforms several extant methods for cross-resolution CD as well as vanilla CD methods in both in-distribution and out-of-distribution settings.
\end{itemize}

% paper structure
The rest of this paper is organized as follows.
Sec. \ref{sec:related-work} introduces related work of existing methods of vanilla CD and those handling bitemporal resolution differences.
Sec. \ref{sec:method} presents the proposed scale-invariant learning with implicit neural networks. 
Experimental results are given in Sec. \ref{sec:experiment}, and the Conclusion is drawn in Sec. \ref{sec:conclusion}.

%  ******************************Related Works****************************** 
\section{Related work}
\label{sec:related-work}

\subsection{Deep Learning-based optical RS Image CD}
The past several years have witnessed remarkable progress in supervised change detection for optical remote sensing imagery using deep learning (DL) techniques. Advanced DL techniques, e.g., CNNs \cite{He2016}, fully convolutional neural networks (FCN) \cite{Long2015}, and transformers \cite{Vaswani2017} have been widely applied in the field of RS CD \cite{Shi2020}.

The predominant recent attempts have aimed to enhance the discriminative capacity of CD models by incorporating advanced network backbones (e.g., HRNet\cite{Liang2022, Cao2020}, vision transformers\cite{Bandara2022a, Zhang2022a, Jiang2022a, Chen2022d}) and network structures (e.g., dilated convolution \cite{Zhang2019c, Lv2022}, deformable convolution \cite{Cheng2022, Song2022b}, attention mechanism \cite{Liu2019b, Zhang2020b, Peng2020a, Raza2022, Lv2022, Li2022e, Chen2020, Feng2022, Li2022c, Song2022, Yuan2022, Liu2022f, Wan2022, Liu2022c, Chen2020e, Zhou2022a, Chen2022, Li2022d, Feng2023}, and flow field-based model \cite{Liu2022c, Fang2022}), devising multi-level bitemporal feature fusion strategies (e.g., UNet \cite{Peng2019, Bem2020, Daudt2018, Hou2020, Zhang2020b, Liu2019b, Papadomanolaki2019, Pei2022, Li2022c, Chen2021a} or FPN\cite{Zhang2020a, Bao2020, Jiang2020}), employing multi-task learning (e.g., additional supervision of land-cover maps for each temporal \cite{Sun2020b, Liu2019b, Papadomanolaki2021, Daudt2019}, boundary supervision of the change edge map \cite{Lei2022, Bai2021, Chen2022d}), combining generative adversarial network (GAN) loss \cite{Hou2020, Zhao2020b}, training with more diverse synthetic data \cite{Chen2021a, Liu2022, Rahman2018}, and fine-tuning from a pre-trained model (e.g., self-supervised pre-training \cite{Manas2021, Wang2022d} and supervised pre-training \cite{Chen2022e}). Note that the paper mainly focuses on binary change detection. The additional supervision of land-cover maps for each temporal \cite{Daudt2019} could also improve the binary change detection performance apart from the purpose of identifying the semantic change categories.

Context modeling, encompassing both spatial context and spatial-temporal context, is crucial for discerning changes of relevance and filtering out extraneous changes across bitemporal images. Among the aforementioned attempts, attention mechanisms, including channel attention\cite{Liu2019b, Zhang2020b, Peng2020a, Raza2022, Lv2022, Li2022e, Chen2020, Fang2021}, spatial attention \cite{Liu2019b, Zhang2020b, Peng2020a, Raza2022, Lv2022}, self-attention \cite{Chen2020e, Chen2020, Li2022d, Zhou2022a, Chen2022}, and cross-attention \cite{Jiang2020, Zhao2020c, Zhou2022, Diakogiannis2020, Feng2023}, have been extensively leveraged as conventional context modeling techniques for the CD task. Early works that incorporate spatial context have primarily focused on employing attention mechanisms as feature enhancement modules, applying them either separately to each temporal image \cite{Liu2019b, Pan2022} or to fused bitemporal features \cite{Zhang2020b, Lv2022, Peng2020a, Raza2022, Li2022e}, lacking the exploit of the temporal-related correlations. More recent works have explicitly modeled spatial-temporal relations by employing cross-attention \cite{Jiang2020, Zhou2022, Diakogiannis2020} or self-attention/transformers on bitemporal features \cite{Chen2022, Feng2022, Song2022, Wang2022a, Shi2022, Wan2022, Song2022b, Song2022c, Liu2022c, Ke2022}. For instance, the Bitemporal Image Transformer (BIT) \cite{Chen2022} applies multi-head self-attention to sparse visual tokens extracted from the per-pixel feature space of bitemporal images to efficiently model their global spatial-temporal relations.

Different from existing context modeling approaches in CD, we introduce local-window self-attention over bitemporal pixels belonging to each small non-overlapping image window. Our motivation stems from the notion that disparities in spatial resolution reflect differences in local textural detail within images. Comparing local regions between bitemporal images of varying resolutions may therefore suffice to align their features. Although Swin Transformer \cite{Liu2021d} whose core is local-window self-attention has been applied in the CD task \cite{Zhang2022a, Yan2022a, Jiang2022a}, it is treated as the mere network backbone, therefore processing bitemporal images independently without modeling their temporal correlations.

Furthermore, most current CD methods have been principally formulated under the assumption of consistent spatial resolution across the bitemporal images. They are thus inadequate for application to cross-resolution paradigms. Our proposed model, in contradistinction, is specifically designed to adapt to resolution differences across bitemporal images.

\subsection{Handling Bitemporal Resolution Differences} 

In light of their real-world applicability, cross-resolution change detection (also termed different-resolution change detection), operable on remote sensing imagery of heterogeneous resolution obtained through different sensors, has claimed burgeoning interest \cite{Tian2022a, Liu2022g, Wang2022j, Tu2017, Zhang2016, Liu2022f, Shao2021}. 
This article mainly focuses on the supervised cross-resolution CD of optical RS images. Works addressing heterogeneous image change detection \cite{Zheng2022a, Zhang2016} for synthetic aperture radar (SAR) and optical data fall outside the purview of this article. 

There exist two predominant categories of cross-resolution CD techniques: those operating in image space and those operating in feature space. 1) Image-space alignment \cite{Tian2022a, Liu2022g, Wang2022j, Tu2017}: first calibrate the spatial scales of bitemporal images and then apply conventional CD methods to the aligned images. The simplest way is to upsample LR images to HR resolution via bilinear/cubic interpolation \cite{Wang2022j} or down-sample HR images to LR ones \cite{Tu2017}. More recently, super-resolution techniques have been deployed for low-to-high-resolution transformation in a learnable fashion \cite{Tian2022a, Liu2022g}. 2) Feature-space alignment: align feature representations via interpolation \cite{Liu2022f, Shao2021}. One Recent work \cite{Liu2022f} applies transformers to learn correlations between the upsampled LR features and original HR ones, achieving semantic alignment across resolutions.

Most existing methods have been formulated solely for scenarios exhibiting a fixed resolution difference, thus inadequate when the resolution discrepancy between bitemporal images varies. Towards more practical real-world applications, we propose a method adaptable to variable resolution differences. Specifically, we learn a scale-invariant embedding space insensitive to changes in resolution via enforcing the model outputs HR CD results regardless of the downsampling factor applied to input HR images. The synthetic reconstructions of randomly downsampled HR images narrow the resolution gap between HR and LR images, thereby achieving adaptability to continuous resolution differences.

\subsection{Implicit Neural Representation}

Implicit Neural Representation (INR), also known as coordinate-based neural representations, is essentially a continuously differentiable function that facilitates transformations from coordinates to signals \cite{Sitzmann2020}. Originally stemming from the field of 3D reconstruction, INR is used to represent the object shape \cite{Genova2019} and 3D scenes \cite{Mildenhall2020a} as a replacement for explicit representations such as point clouds, meshes, or voxels. Thanks to the design of coordinate-based representation, INR exhibits an ability to model images of variant resolutions, thus being employed in image processing tasks such as super-resolution \cite{Chen2021c, Xu2021b}, semantic segmentation \cite{Shen2022, Hu2022}, and instance segmentation \cite{Cheng2022a}.

Recently, INR has been applied in the field of RS \cite{Wu2022a, Qi2023, Qi2022, Liu2023, Chen2023, Luo2023, Chen2023a}, including 3D RS scene reconstruction \cite{Wu2022a} and segmentation \cite{Qi2023, Qi2022}, 2D RS image synthesis \cite{Liu2023}, and super-resolution \cite{Chen2023, Luo2023}. However, the employment of INR for RS 2D image understanding remains limited, particularly for the task of CD which has received scarce exploration. For the task of cross-resolution CD, we incorporate INR to enhance model adaptability to cross-temporal resolution differences. Our motivation is that the INR may enable the implicit encoding of the shape of change objects and extraction of the corresponding HR change mask from latent space based on coordinate queries, regardless of the resolution difference across bitemporal images.

%  ******************************Methods****************************** 

\section{Scale-invariant Learning with Implicit Neural Representations for Cross-resolution Change Detection}
\label{sec:method}

In this section, we first give an overview of the proposed scale-invariant cross-resolution method and then introduce its three main components. Finally, implementation details are given.

\subsection{Overview}

Cross-resolution change detection aims to obtain an HR change mask based on bitemporal images with different resolutions (i.e., an LR image and an HR image). Towards real-world applications, here we propose Scale-Invariant Learning with Implicit neural networks (SILI) for handling varying resolution difference ratios across bitemporal images. 

The essence of the proposed method is to learn a scale-invariant embedding space regardless of the resolution discrepancies between bitemporal images, and decode the high-resolution change mask with dense coordinate queries and corresponding multi-level features by leveraging the implicit neural representations encapsulating the shape of local changes.

Fig. \ref{fig:overall} illustrates the proposed SILI. It has three main components, including random resolution image synthesis, image encoder with feature-level bitemporal interaction, and change decoder based on implicit neural representations.

1) \textbf{Random resolution image synthesis}. Rather than manipulating the original bitemporal images with a fixed resolution difference ratio, we perform random downscaling reconstruction on the HR image (i.e., downsampling succeeded by upsampling) to narrow the resolution gap between HR and LR images. Moreover, we propose random bitemporal region swapping to further improve the model adaptability to scale variance. For more details see Sec. \ref{subsec:synthesis}.

2) \textbf{Image encoder}. Given synthesized bitemporal images, a normal Siamese CNN backbone (e.g., ResNet-18 \cite{He2016}) is employed to obtain multi-level image features for each temporal. Bitemporal features of certain levels are interacted with each other by leveraging local-window self-attention to reduce the semantic gap caused by resolution differences. Details of the bitemporal feature interaction see Sec. \ref{subsec:encoder}

3) \textbf{Change decoder}. Instead of upsampling multi-level bitemporal features to the target size with traditional interpolation, we incorporate coordinate-based representations to decode the label for each position by feeding corresponding multi-level features and position embeddings to a multilayer perceptron (MLP) that implicitly learns the shape of local changes. See Sec. \ref{subsec:decoder} for more details.

\subsection{Random Resolution Image Synthesis}
\label{subsec:synthesis}

\begin{figure*}
        \centering
        \includegraphics[width=0.9\textwidth]{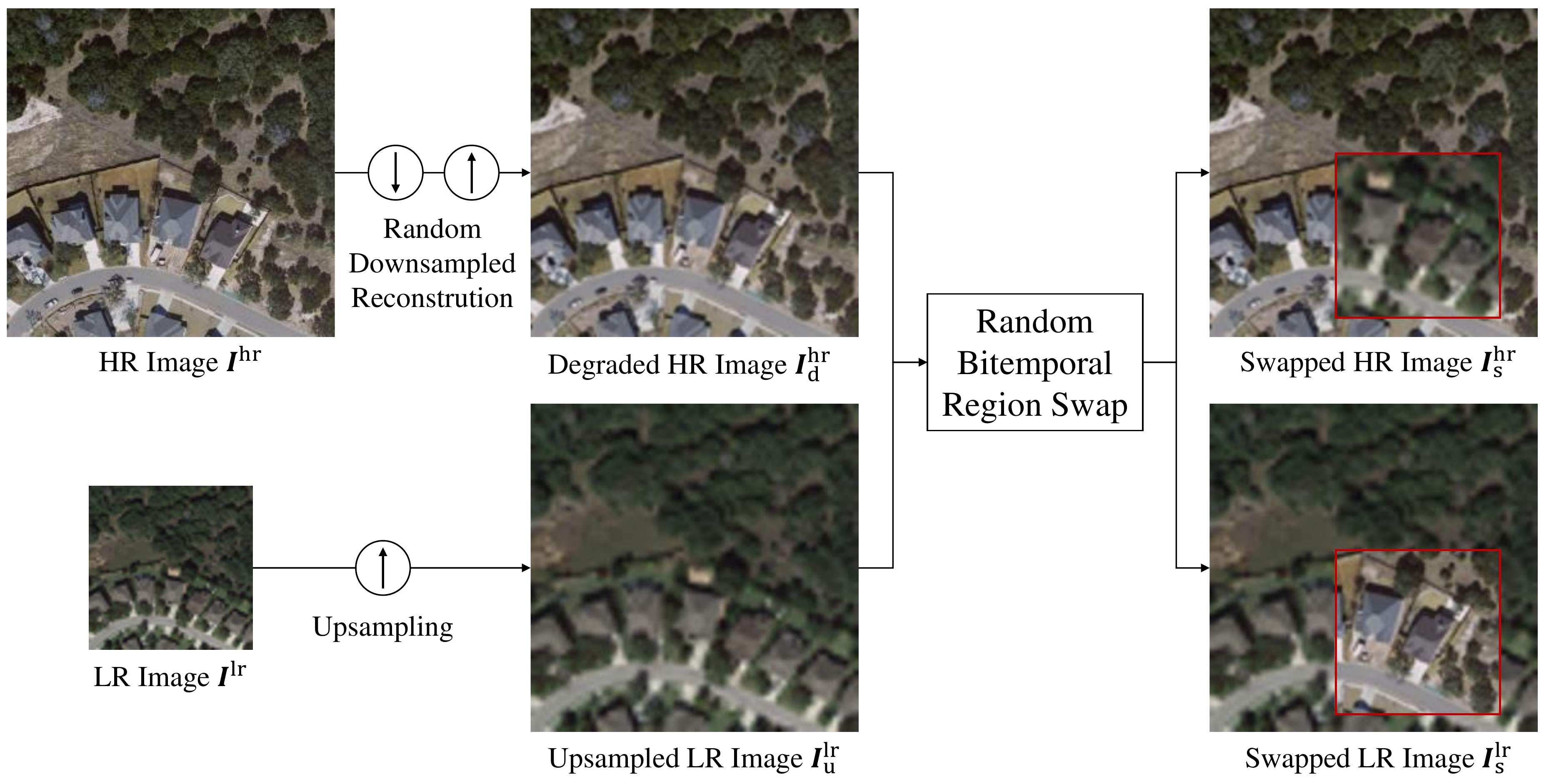}
        \caption{Illustration of random resolution image synthesis, including 1) upsampling the LR image, 2) random downsampling and reconstructing the HR image, and 3) random region swap between the upsampled LR image and the degraded HR image.
        }
        \label{fig:random_syn}
\end{figure*}

In the training phase, we introduce scale-invariant learning by compelling the change detection model to generate HR change masks for synthesized bitemporal images subject to random scale manipulation, thus enhancing the adaptation capacity for handling continuous resolution difference ratios across bitemporal images. Specifically, given bitemporal images ($\boldsymbol{I}^{\text{lr}} \in \mathbb{R}^{H_{\text{lr}} \times W_{\text{lr}} \times 3}, \boldsymbol{I}^{\text{hr}} \in \mathbb{R}^{H_{\text{hr}} \times W_{\text{hr}} \times 3}$) with resolution differences ratio ($r_d=H_{\text{hr}}/H_{\text{lr}}$), we design three main steps: 1) upsampling the LR image to the same size as that of the HR image, 2) perform random downsampling reconstruction on the HR image, 3) random region swap between bitemporal images. Note that the resolution differences ratio defines as the ratio of the ground resolution of the LR image and that of the HR image. For example, $r_d=4$ for bitemporal images with ground resolution 0.5m/pixel and 2m/pixel. Fig. \ref{fig:random_syn} illustrates the overall process of random resolution image synthesis.
% 可以加一段算法描述；

1) \textbf{Upsampling LR image}. We first upsample the LR image $\boldsymbol{I}^{\text{lr}}$ to the size of $\boldsymbol{I}^{\text{hr}}$ via bicubic interpolation. Instead of learning upsampled LR images via training a super-resolution reconstruction model, we aim to learn a scale-invariant CD model that is able to handle degraded constructions with essentially lower resolutions than the HR image. Formally, The upsampled LR image $\boldsymbol{I}^{\text{lr}}_{u} \in \mathbb{R}^{H_{\text{hr}} \times \times W_{\text{hr}} \times 3}$ is given by 
\begin{equation}
    \boldsymbol{I}^{\text{lr}}_{u} = \text{upsampling}({I}^{\text{lr}}, r_d).
\end{equation}

2) \textbf{Random downsampled reconstruction}. To acclimatize the model to various resolution differences, we synthesize degraded variants of the HR image through downsampling by a random ratio, thereafter rescaling to the initial size. Formally, we randomly sample a ratio $r$ from the Uniform distribution $r\sim U[1, r_d]$. The downsampled reconstruction version $\boldsymbol{I}^{\text{hr}}_{d} \in \mathbb{R}^{H_{\text{hr}} \times W_{\text{hr}} \times 3}$ can be given by 
\begin{equation}
    \boldsymbol{I}^{\text{hr}}_{d} = \text{upsampling}(\text{downsampling}(\boldsymbol{I}^{\text{hr}}, r), r),
\end{equation}
where bicubic interpolation is applied to implement upsample and downsample.

3) \textbf{Random bitemporal region swap}. Considering that simply downsampling the HR image may not wholely fill the gap between the HR and the LR image captured by different sensors, we further propose to swap a randomly selected region between bitemporal images. Such operation can be viewed as a form of image-level bitemporal interaction, allowing the CD model to process LR and HR data concurrently, which may benefit learning more scale-invariant representations. Formally, the swapped bitemporal images $\boldsymbol{I}^{\text{hr}}_{s}, \boldsymbol{I}^{\text{lr}}_{s} \in \mathbb{R}^{H_{\text{hr}} \times W_{\text{hr}} \times 3}$ are given by 
\begin{equation}
    \boldsymbol{I}^{\text{hr}}_{s}, \boldsymbol{I}^{\text{lr}}_{s} = \text{swap}(\boldsymbol{I}^{\text{hr}}_{d}, \boldsymbol{I}^{\text{lr}}_{u}, u, v, \text{crop\_size}),
\end{equation}
where $(u, v), u \sim U[1, W_{hr}-\text{crop\_size}], v \sim U[1, H_{hr}-\text{crop\_size}]$ is the coordinate of the upper-left point of the cropped region and $\text{crop\_size}$ is the size of the swapped region. $\text{crop\_size}$ is default set to half of $W_{hr}$. 

Note that in the inference/testing phase, we only perform the first step, i.e., rescale the LR image to the size of the HR image. In other words, we do not apply the random downsampled reconstruction and random bitemporal region swap in the testing phase.

\subsection{Image Encoder with Bitemporal Local Interaction}
\label{subsec:encoder}
Given synthesized bitemporal images, we employ an off-the-shell Siamese CNN backbone (i.e., ResNet-18) for generating multi-level features $\boldsymbol{X}^{i}_{j} \in \mathbb{R}^{H_{j} \times W_{j} \times C_{j}}$ for each temporal image $i\in \{1,2\}$. Note that $j\in \{1, 2, 3, 4\}$ denotes the level of the generated features with the size of $H_{j} \times W_{j}, H_{j}=H_{\text{hr}}/2^{(j+1)}, W_{j}=W_{\text{hr}}/2^{(j+1)}$. Instead of encoding bitemporal images independently without interaction, we supplement feature interaction between bitemporal image features of a certain level to refine them via modeling local spatial-temporal correlations thus benefiting feature extraction at the next level. 

Concretely, we introduce local-window self-attention \cite{Liu2021d} over bitemporal pixels within each non-overlapping window on the feature map of a certain level. Our incentive resides in that discrepancies in spatial resolution between images may predominantly mirror local texture variances in land cover and thus leveraging local correlations may in turn benefit aligning features of bitemporal images with different resolutions. 

\begin{figure}
        \centering
        \includegraphics[width=0.5\textwidth]{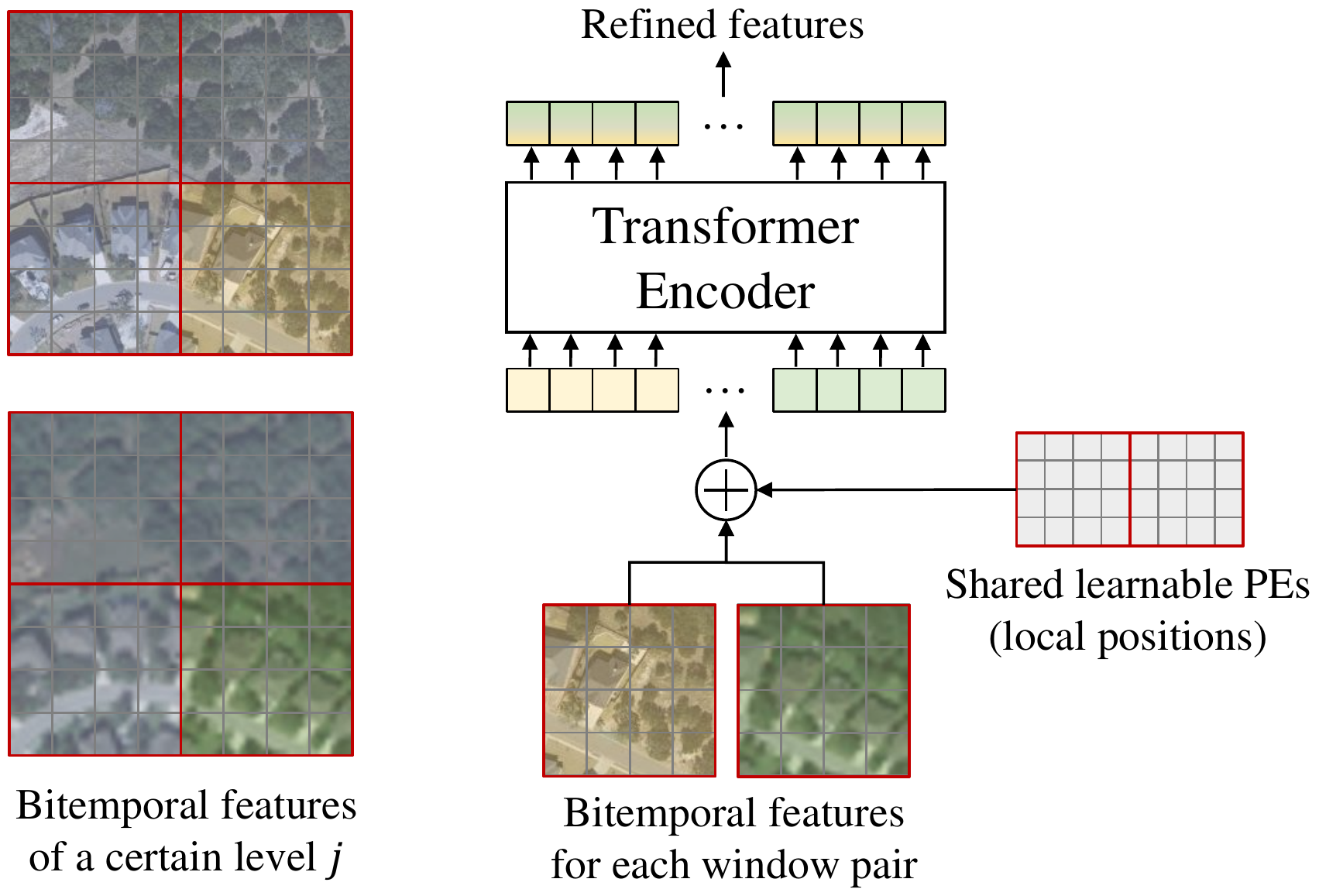}   
        \caption{Illustration of bitemporal interaction based on local-window self-attention. For bitemporal features of a certain level $j$, we perform multi-head self-attention within each window pair to better semantically align features from different resolution images. 
        }
        \label{fig:local_att}
\end{figure}

Fig. \ref{fig:local_att} illustrates the proposed bitemporal interaction based on local-window self-attention.
For bitemporal features $\boldsymbol{X}^{1}_{j}, \boldsymbol{X}^{2}_{j}$ of a certain level $j$, we evenly partition them into non-overlapped windows. Let $\boldsymbol{X}^{1}_{j, n}, \boldsymbol{X}^{2}_{j, n} \in \mathbb{R}^{\text{WS} \times \text{WS} \times C_{j}}, n \in \{1,...,N_{w}\}$ be bitemporal features within each window, where $\text{WS}$ is the window size, $n$ denotes the window index in $N_{w}$ partitioned windows. We apply multi-head self-attention (MSA) on bitemporal patches within each local window. Formally, the refined bitemporal features $\boldsymbol{X}^{1*}_{j,n}, \boldsymbol{X}^{2*}_{j,n}$ of level $j$ are given by
\begin{equation}
    \boldsymbol{X}^{1*}_{j,n}, \boldsymbol{X}^{2*}_{j,n} = \text{Transformer\_Encoder}(\boldsymbol{X}^{1}_{j,n}, \boldsymbol{X}^{2}_{j,n}),
\end{equation}
where a vanilla transformer encoder \cite{Vaswani2017} is employed to implement multi-head self-attention. Note that we apply shared learnable local positional embeddings (PE) \cite{Vaswani2017} for each window. The PE could encode temporal and local spatial position information, thus helping model spatial-temporal relations. The transformer encoder consists of $N_{te}$ transformer layers. Our empirical evidence (see Sec. \ref{subsec:parameter}) suggests local context modeling at early layers is sufficient. Please refer to \cite{Vaswani2017} for more details on the transformer layer. 
Note that the calculation of local-window self-attention for each window could be processed in parallel in GPU.

After gleaning bitemporal multi-level features, we further transform the output features to a uniform dimension $C$ by applying one $1\times 1$ convolution layer to each level. The transformed bitemporal image features $\boldsymbol{Z}^{1}_{j}, \boldsymbol{Z}^{2}_{j} \in \mathbb{R}^{H_{\text{hr}} \times \times W_{\text{hr}} \times 2C}$ of each level $j$ are then fused via channel-wise concatenation. The resulting multi-level features $\{\boldsymbol{Z}_{j}| j\in \{1,2,3,4\} \}$ are given by
\begin{equation}
    \boldsymbol{Z}_{j} = \text{Concat}(\boldsymbol{Z}^{1}_{j}, \boldsymbol{Z}^{2}_{j}).
\end{equation}

Apart from the multi-level features from the vanilla backbone, we extract handcrafted low-level edges from bitemporal images as spatial clues to obtain high-resolution change masks in the subsequent change decoder. It is motivated by the evidence \cite{zheng2022hfa, shangguan2023contour, liu2022idan} that the incorporation of handcrafted edge features (e.g., Canny \cite{Canny1986}, Sobel, or Prewitt operator) within the deep neural networks benefits the change detection performance. As the Canny operator could obtain more clean and accurate edges than the Sobel operator, we chose the Canny operator to extract low-level edge clues. Here, we simply utilize the Canny operator on each dimension of bitemporal images to obtain handcrafted edge features which are then fed into the change decoder. Formally, the edge features $\boldsymbol{X}_{0} \in \mathbb{R}^{H_{\text{hr}} \times \times W_{\text{hr}} \times 3}$ are given by the channel-wise summation of that from each temporal image as follows:
\begin{equation}
    \boldsymbol{X}_{0} = \text{Canny}(\boldsymbol{I}^{\text{hr}}_{s}) + \text{Canny}(\boldsymbol{I}^{\text{lr}}_{u}).
\end{equation}
Note that for simplicity, we directly perform pixel-wise addition of bitemporal Canny features to obtain the edge clues.

Similarly, the handcrafted edge $\boldsymbol{X}_{0}$ goes through a relatively large kernel ($7\times 7$) convolution layer to obtain the learned edge clues $\boldsymbol{Z}_{0} \in \mathbb{R}^{H_{\text{hr}} \times \times W_{\text{hr}} \times 3}$ that are then fed into the subsequent change decoder.

\subsection{Change Decoder with Implicit Neural Representation}
\label{subsec:decoder}
Given multi-level bitemporal image features and edge clues, we aim to decode the HR change mask $CM \in \mathbb{R}^{H_{\text{hr}} \times W_{\text{hr}}}$ by leveraging implicit neural representation (INR), viz. feeding dense coordinates alongside corresponding image features to a learnable MLP that implicitly represents the shape of local changes. Our motivation is that the INR may assist in reconstructing the detailed shape of the degraded land cover of change from the LR image by leveraging fine features from HR images. The key is to learn implicit neural networks $f_{\theta}$ (typically an MLP) over coarse resolution feature maps to define continuous representations that could yield the HR change mask according to the coordinate queries of the HR grid.

Now, we define a continuous normalized 2D space $\mathbb{S}=\{x=(u,v) | u, v \in [0,1]\}$. Images or feature maps of different levels can be evenly distributed in the space $\mathbb{S}$ where each cell in the grid is assigned a 2D coordinate according to its center position. For instance, given a position indexed $h$-th, $w$-th ($h \in \{0,1,...,H-1\}, w \in \{0,1,...,W-1\}$) in an grid of size $H \times W$, its coordinate in space $\mathbb{S}$ is $(u,v) =(\frac{1}{2H}+\frac{h}{H}, \frac{1}{2W}+\frac{w}{W})$.

\begin{figure}
        \centering
        \includegraphics[width=0.5\textwidth]{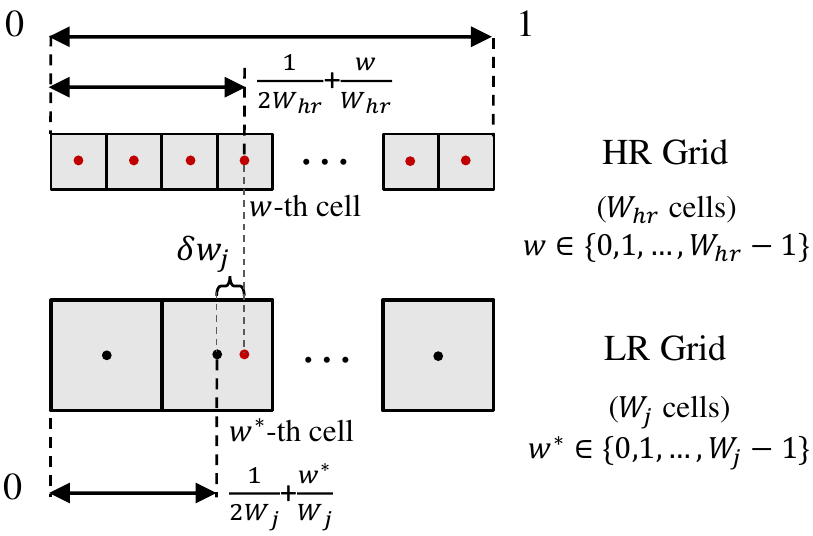}   
        \caption{Illustration of grid coordinates in the normalized space $\mathbb{S}$. Here, the HR grid denotes the dense coordinate queries and the LR grid denotes the feature map from a certain level $j$. Only the width direction is shown for a better view. We also demonstrate the coordinate matching between the query ($w$-th) cell in the HR grid and its nearest ($w^{*}$-th) cell in the LR grid.  
        }
        \label{fig:coordinate}
\end{figure}

Fig. \ref{fig:coordinate} illustrates the coordinate relations between the HR grid and a relatively LR grid with respect to the space $\mathbb{S}$. We only show one dimension (width direction) for a better view. Here, the HR grid denotes the dense coordinate queries while the LR grid denotes the feature map from a certain level.

\textbf{Query features from relatively LR feature maps}. Let $x_{q}$ be the coordinate of point $q$ in an HR grid with respect to $\mathbb{S}$. Given one query $x_{q}$, we need first to collect corresponding features on the coarse feature map $\boldsymbol{Z}_{j}$ by calculating the nearest cell to the query for each level $j\in \{1,2,3,4\}$. The matched coordinate $x^{*}_{q,j}$ for $\boldsymbol{Z}_{j}$ can be given by 
\begin{equation}
    x^{*}_{q,j} = (\frac{1}{2H_{j}}+\frac{h^{*}}{H_{j}}, \frac{1}{2W_{j}}+\frac{w^{*}}{W_{j}}),
\end{equation}
where $(h^{*}, w^{*})$ is the corresponding coordinate of the matched point in the discrete space of $\boldsymbol{Z}_{j}$. $h^{*}, w^{*}$ can be calculated as follows:
\begin{align}
        h^{*} &= \text{round}(\frac{H_{j}}{H_{hr}}(\frac{1}{2}+h)-\frac{1}{2}) \\
        w^{*} &= \text{round}(\frac{W_{j}}{W_{hr}}(\frac{1}{2}+w)-\frac{1}{2}),
\end{align}
where $h, w$ is the height/width index of $q$ in the HR image, respectively. 

Because the feature map of level $0$ has the same resolution as the HR coordinate grid, the corresponding feature vector $\boldsymbol{Z}_{0}[x_{q}]$ can be directly obtained by the query coordinate $x_{q}$. Note that for a tradeoff between accuracy and efficiency (see \ref{subsec:parameter}), the input coordinate grid is downsampled by a factor of 2 compared to the original HR image.

\textbf{Relative positional encoding (PE)}. We further calculate the relative coordinate encoding $\text{PE}_{q,j} \in \mathbb{R}^{C_{\text{pe}}}$ for level $j$ between $x_{q}$ and $x^{*}_{q,j}$:
\begin{equation}
    \text{PE}_{q,j} = \phi(\delta x_{q,j}) = \phi(x_{q} - x^{*}_{q,j}), 
\end{equation}
where $\phi(\cdot)$ denotes the position encoding function \cite{Vaswani2017} to transform the 2D coordinate to high dimensional vectors that are more capable of representing high-frequency signals. $\delta x_{q,j}$ is the relative coordinate between the query and its nearest grid cell center.

\textbf{Encode cell scale}. Considering that the grid cell of different resolutions occupies different spatial scopes, to distinguish features from different levels, we also combine cell scale, i.e., the absolute height and width of a cell with respect to the continuous space $\mathbb{S}$ as follows:
\begin{equation}
    \text{CS}_{q,j} = [\frac{1}{H_{j}}, \frac{1}{W_{j}}],
\end{equation}
where $\text{CS}_{q,j} \in \mathbb{R}^{2}$ is the cell scale for feature level $j, j\in \{1,2,3,4\}$.

\textbf{Decode change probability}. As shown in Fig. \ref{fig:overall}, after obtaining multi-level features and corresponding PEs and cell scales, an MLP is employed to decode the change probability for each query as follows:
\begin{equation}
    \boldsymbol{P}[x_{q}] = f_{\theta}(\text{Concat}(\boldsymbol{Z}_{0}[x_{q}], \{\boldsymbol{Z}_{j}[x^{*}_{q,j}], \text{PE}_{q,j}, \text{CS}_{q,j}\}_{j=1}^{4})),
\end{equation}
where $\text{Concat}(\cdot)$ denotes to channel-wise concatenate the input items. $\boldsymbol{P} \in \mathbb{R}^{H_{\text{hr}} \times \times W_{\text{hr}} \times 2}$ is the change score maps where the 2D vector for each position indicates the probability of change or not.

\subsection{Implementation Details}
\label{subsec:details}
\textbf{CNN backbone}. We employ the off-the-shell ResNet-18 as the CNN backbone. Its intermediate multi-level features (channel dimensions $C_{j}$ are 64, 128, 256, 512, respectively for level $j \in \{1,2,3,4\}$) are transformed to the same dimension $C=64$ via one convolution layer. We apply the bitemporal interaction with a local window size $\text{WS}=8$ at level $j=1,2,3$.

\textbf{Change decoder}. The channel dimension ${C_{\text{pe}}}$ of the relative positional encoding is set to 24. The implicit neural network $f_{\theta}$ is implemented by a three-layer MLP with BatchNorm and ReLU in between. The output channel dimension of each MLP is "64, 64, 2", respectively.  

\textbf{Loss function}. 
In the training phase, we minimize the cross-entropy loss between the predicted change probability map $\boldsymbol{P}$ and ground truth to optimize the network parameters. In the inference phase, the change mask can be obtained by per-pixel Argmax operation on the channel dimension of $\boldsymbol{P}$.

%  ******************************Experiments****************************** 
\section{Experimental Results}
\label{sec:experiment}

\subsection{Experimental Setup}

To evaluate the proposed cross-resolution CD model, we conduct experiments on two synthesized cross-resolution CD datasets (construct the LR image by downsampling), and one real-world cross-resolution CD dataset (the LR and HR image pair captured by different-resolution satellite sensors).

\textbf{LEVIR-CD(4$\times$)}. LEVIR-CD \cite{Chen2020e} is a widely used building CD dataset, which contains 637 pairs of bitemporal VHR (0.5m/pixel) images, each size of $1024\times 1024$. We follow the default dataset split \cite{Chen2020e}, including 445/64/128 samples for training/validation/testing, respectively. We further crop each sample into small patches of size $256\times 256$ with no overlap. To synthesize cross-resolution scenarios, we downsample the post-event (t2) image for each sample by a ratio of 4. In this way, we obtain the simulated LEVIR-CD(4$\times$), where the post-event image has a 4 times spatial resolution lower than that of the pre-event (t1) image.

\textbf{SV-CD(8$\times$)}. Season-varying change detection (SV-CD) \cite{Lebedev2018} is another widely used binary CD dataset. It contains 11 pairs of VHR (0.03\~1m/pixel) RGB images with sizes ranging from $1900 \times 1000$ to $4725 \times 2700$ pixels. The changes in buildings, cars, and roads are taken into consideration. We follow the default dataset split, which contains 10000/3000/3000 cropped samples each size of $256\times 256$ for training/validating/testing, respectively. For each sample, we downsample the post-event image by a ratio of 8, thus obtaining the synthesized SV-CD(8$\times$).

\textbf{DE-CD(3.3$\times$)}. DynamicEarthNet \cite{Toker2022} is a multi-class land use and land cover (LULC) segmentation and change detection dataset for daily monitoring of Earth's surface. It covers 75 areas of interest (AOIs) around the world and consists of samples captured in the range from 2018-01-01 to 2019-12-31. Each AOI provides high-time-frequency (daily) Planet imagery (3m/pixel) and monthly LULC per-pixel annotations, as well as monthly Sentinel-2 imagery (upsampled to match the size of the Planet image) whose original spatial resolution is 10m/pixel. Each sample has a size of $1024\times 1024$. We reorganize the original dataset for cross-resolution change detection. We collect the time-aligned (monthly) image and label data, where the Sentinel-2 data in each month in 2018 is for pre-event and the Planet data captured 1 year later than the Sentinel-2 is for post-event. For simplicity, we only focus on the changes in the land cover belonging to impervious surfaces. We exclude those without changes of interest and therefore obtain 506 samples, which are then randomly split into 354/51/101 samples for training/validating/testing. In this way, we have aggregated DE-CD(3.3$\times$) where the bitemporal resolution difference ratio is around 3.3. Similarly, we cropped each sample into $256\times 256$ small patches with no overlap.

To evaluate the proposed method, we set the following models for comparison:

1) \textbf{Base}. Our baseline consists of a CNN backbone (ResNet-18) and a change decoder with the channel-wise concatenated input of bitemporal transformed features (channel dimension of $64$) at each level ($j\in \{1,2,3,4\}$) from the encoder. Similar to our INR decoder, the baseline decoder has three-layer convolutions (output dimensions of 64,64,2) with BatchNorm and ReLU in between.

2) \textbf{SILI}. Our proposed SILI model with the random resolution image synthesis, a ResNet-18-based encoder with bitemporal feature interactions, and a change decoder with INR.

\textbf{Training details}. Data augmentation techniques, including random flip, and Gaussian blur are applied. We employ SGD with a batch size of 8, a momentum of 0.9, and a weight decay of 0.0005. The initial learning rate is 0.01 and linearly decays to 0 until 200 epochs. We evaluate the model using the validating set at the end of each training epoch. The best validating model is evaluated on the test set.

\textbf{Evaluation Metrics}. We use the F1-score regarding the change category as the evaluation metrics. Precision, recall, and Intersection over Union (IoU) belonging to the change category are also reported. These indices can be defined by: 
\begin{equation}
\begin{split}
        & \text{F1} = \frac{2}{\text{recall}^{-1}+\text {precision}^{-1}} \\
        & \text{precision} = \frac{\text{TP}}{\text{TP} + \text{FP}} \\
        & \text{recall} = \frac{\text{TP}}{\text{TP} + \text{FN}} \\
        & \text{IoU} = \frac{\text{TP}}{\text{TP} + \text{FN} + \text{FP}} \\
\end{split}
\end{equation}
where $\text{TP, FP, FN}$ are the number of true positives, false positives, and false negatives, respectively.

\subsection{Overall Comparison}
\label{ssec:comparison}

We make a comparison with several state-of-the-art conventional change detection methods, including three pure convolutional-based methods (FC-EF \cite{Daudt2018}, FC-Siam-Diff \cite{Daudt2018}, FC-Siam-Conc \cite{Daudt2018}), and six attention-based methods (STANet \cite{Chen2020e}, IFNet \cite{Zhang2020b}, IFNet \cite{Zhang2020b}, SNUNet \cite{Fang2021}, BIT \cite{Chen2022}, ICIFNet \cite{Feng2022}, DMINet \cite{Feng2023}). We also compare two CD methods (SUNet \cite{Shao2021}, SRCDNet \cite{Liu2022g}) specifically for the scenario of different resolutions across bitemporal images. 
\begin{itemize}
    \item FC-EF \cite{Daudt2018}. Image-level fusion method where bitemporal images are channel-wise concatenated to be fed into an FCN.
    \item FC-Siam-Diff \cite{Daudt2018}. Feature-level fusion method where a Siamese FCN is employed to obtain multi-level features for each temporal image, then bitemporal feature differencing is calculated for fusing temporal information.
    \item FC-Siam-Conc \cite{Daudt2018}. Feature-level fusion method where channel-wise concatenation is used for fusing temporal information.
    \item STANet \cite{Chen2020e}. Metric-based method, which incorporates multi-scale self-attention to enhance the discriminative capacity for bitemporal features.
    \item IFNet \cite{Zhang2020b}. Feature-level concatenation method, which employs channel/spatial attention on the concatenated bitemporal features at each level of the decoder. Deep supervision is applied on each level for better training of the intermediate layers.  
    \item SNUNet \cite{Fang2021}. Feature-level concatenation method, which employs NestedUNet \cite{Zhou2018} to extract multi-level bitemporal features. Channel attention and deep supervision are applied on each level of the decoder.
    \item BIT \cite{Chen2022}. Feature-level differencing method, which expresses the input images into a few visual words (tokens), and models spatiotemporal context in the token-based space by transformers to efficiently benefit per-pixel representations.  
    \item ICIFNet \cite{Feng2022}. Feature-level differencing method, which integrates CNN and Transformer to parallelly extract multi-level bitemporal features. Cross-attention is applied to fuse parallel features at each level.
    \item DMINet \cite{Feng2023}. Feature-level fusion method, which combines self-attention and cross-attention on bitemporal features of each level to perform temporal interactions, and uses both feature differencing and concatenation parallelly to obtain the change information. Deep supervision is also applied for better performance.
    \item SUNet \cite{Shao2021}. Feature-space alignment method, which designs an asymmetric convolutional network in the early stage of the encoder to spatially align HR/LR images. Handcrafted edge maps for each bitemporal image are also fed into the model as auxiliary information. For a fair comparison, we implement it by upsampling the LR image to the size of the HR image to eliminate the loss of small targets.
    \item SRCDNet \cite{Liu2022g}. Image-space alignment method, which jointly optimizes a GAN-based image super-resolution model and a change detection model. For a fair comparison, due to the inaccessibility of the ground truth HR version for the LR image, we use the pair of HR images and their downsampled version to train the super-resolution model and apply it to the LR image to obtain the upsampled LR image in the inference phase.
\end{itemize}

We implement the above change detection models using their public codes with default hyperparameters. Note that for adapting these conventional CD methods to the cross-resolution CD task, we resize the LR image to the size of the HR image by cubic interpolation before feeding them into the CD model.

Table \ref{tab:comparison_sota} reports the overall comparison results on the LEVIR-CD(4$\times$), SV-CD(8$\times$), and DE-CD(3.3$\times$) test sets. In this setting, each compared model is tested by the bitemporal samples with fixed resolution difference ratios the same as in the training phase. Quantitative results show that our proposed method consistently outperforms the compared conventional CD methods as well as cross-resolution CD methods in terms of F1/IoU/OA scores across the three datasets. Note that as the pure convolutional-based methods (FC-EF, FC-Siam-Conc, and FC-Siam-Diff) fail to fit the DE-CD(3.3$\times$) training set, therefore their performance scores are omitted.

\begin{table*}
    \centering
    \caption{Comparison results on the three CD test sets. The best results are marked in \textbf{bold}. All the scores are described as percentages (\%). }
    \resizebox{1\textwidth}{!}{
    \begin{tabular}{c|c|c|c}
  \toprule
    \multicolumn{1}{c}{} &
    \multicolumn{1}{|c|}{\textbf{LEVIR-CD(4$\times$)}}  &
    \multicolumn{1}{c|}{\textbf{SV-CD(8$\times$)}}  &
    \multicolumn{1}{c}{\textbf{DE-CD(3.3$\times$)}}  \\
    & Pre. / Rec. / F1 / IoU / OA & Pre. /  Rec. / F1 / IoU / OA & Pre. /  Rec. / F1 / IoU / OA \\
    \midrule
    FC-EF \cite{Daudt2018} 
    & 79.57 / 71.48 / 75.31 / 60.40 / 97.64
    & 74.25 / 45.03 / 56.06 / 38.95 / 91.67
    & - \\
    FC-Siam-Conc \cite{Daudt2018} 
    & 84.23 / 69.90 / 76.40 / 61.81 / 97.82  
    & 73.11 / 50.87 / 59.99 / 42.85 / 92.00 
    & -\\
    FC-Siam-Diff \cite{Daudt2018} 
    & 86.12 / 60.15 / 70.83 / 54.83 / 97.50  
    & 76.10 / 56.68 / 64.97 / 48.12 / 92.79 
    & - \\
    STANet \cite{Chen2020e} 
    & 57.87 / 45.47 / 50.93 / 34.16 / 95.58
    & 83.06 / 70.73 / 76.41 / 61.82 / 94.85 
    & 11.80 / 46.89 / 18.85 / 10.41 / 97.61\\
    IFNet \cite{Zhang2020b}  
    & 86.81 / 80.85 / 83.73 / 72.01 / 98.41
    & 94.94 / 79.62 / 86.61 / 76.38 / 97.09 
    & 26.20 / 52.64 / 34.98 / 21.20 / 98.84\\
    SNUNet \cite{Fang2021} 
    & 89.67 / 81.00 / 85.11 / 74.09 / 98.57
    & 92.61 / 83.80 / 87.98 / 78.55 / 97.30 
    & 38.21 / 37.16 / 37.68 / 23.21 / 99.27 \\
    BIT \cite{Chen2022}  
    & 89.57 / 82.11 / 85.68 / 74.94 / 98.61
    & 97.09 / 84.80 / 90.53 / 82.69 / 97.91 
    & 62.05 / 33.38 / 43.41 / 27.72 / 99.48\\
    ICIFNet \cite{Feng2022}  
    & 87.84 / 84.62 / 86.20 / 75.75 / 98.63 
    & 95.68 / 90.56 / 93.05 / 87.00 / 98.40 
    & 63.50 / 25.04 / 35.92 / 21.89 / 99.47 \\
    DMINet \cite{Feng2023}  
    & 89.66 / 84.28 / 86.89 / 76.82 / 98.72
    & \textbf{97.77} / 89.76 / 93.60 / 87.96 / 98.55
    & \textbf{71.47} / 33.84 / 45.93 / 29.81 / 99.53 \\
    \midrule
    SUNet \cite{Shao2021}  
    & 64.12 / \textbf{93.54} / 76.08 / 61.40 / 97.03
    & 63.55 / \textbf{97.98} / 77.10 / 62.73 / 93.13
    & 32.60 / \textbf{71.00} / 44.68 / 28.77 / 98.96 \\
    SRCDNet \cite{Liu2022g}  
    & 66.29 / 84.18 / 74.17 / 58.94 / 97.04
    & 91.30 / 91.89 / 91.59 / 84.49 / 98.01 
    & 39.62 / 33.22 / 36.13 / 22.05 / 99.30 \\
    \midrule
    Base
    & 89.56 / 84.24 / 86.81 / 76.70 / 98.71
    & 96.11 / 89.00 / 92.42 / 85.90 / 98.28 
    & 58.50 / 27.38 / 37.30 / 22.93 / 99.45\\
    Ours & \textbf{90.55} / 86.30 / \textbf{88.38} / \textbf{79.18} / \textbf{98.86}
    & 95.29 / 93.36 / \textbf{94.32} / \textbf{89.24} / \textbf{98.67} 
    & 61.35 / 42.32 / \textbf{50.10} / \textbf{33.42} / \textbf{99.50}\\
   \bottomrule
    \end{tabular}
    }
    \label{tab:comparison_sota}
\end{table*}

\begin{figure*}
        \centering
        \includegraphics[width=1\textwidth]{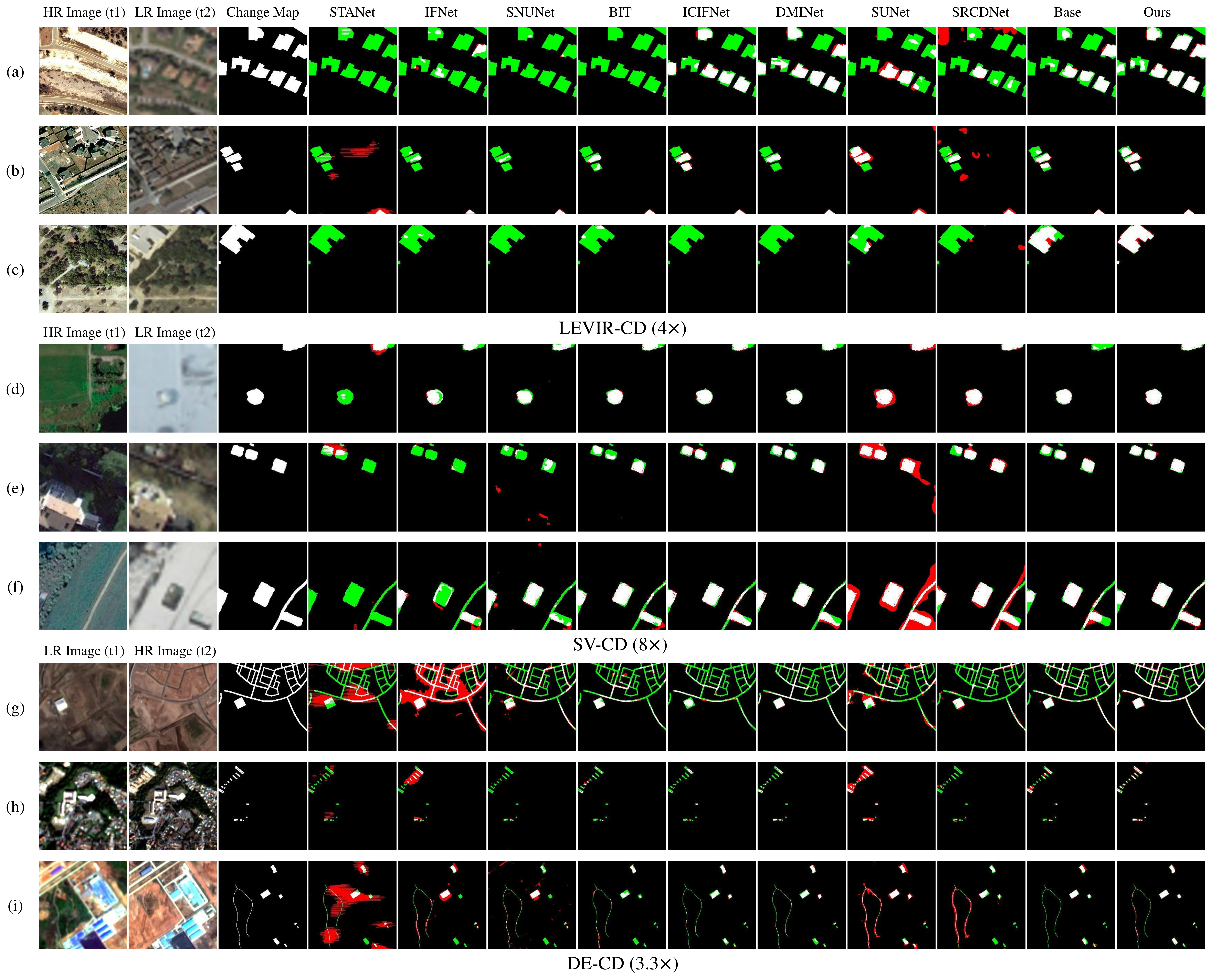}
        \caption{Visual results of the compared methods on the three datasets. For a better view, we use white for true positive, black for true negative, red for false positive, and green for false negative.
        }
        \label{fig:comparison_sota}
\end{figure*}

\textbf{Comparison with conventional CD methods}. We can observe from Table \ref{tab:comparison_sota} that the conventional change detection models with feeding image-space aligned bitemporal inputs by interpolating LR images to the size of HR images can be viewed as strong counterparts in the cross-resolution setting. For example, the recent transformer-based methods (e.g., BIT and ICIFNet) could yield competitive even superior performance over specially designed cross-resolution CD models (SUNet and SRCDNet). It indicates that state-of-the-art conventional CD models can be effectively adapted to the cross-resolution change detection task via simple interpolation-based image-level alignment. Despite the common design of the model structure without sophisticated multi-scale feature fusion strategies (e.g., UNet-based incremental aggregation \cite{Fang2021, Zhang2020b, Feng2023}), or transformer structures \cite{Chen2022, Feng2022}, our proposed method with the MLP-based change decoder could surpass extant methods.

\textbf{Comparison with cross-resolution CD methods}. Quantitative results show that our proposed method significantly precedes the compared cross-resolution methods on the three datasets. Worth noting that our baseline is comparable or even superior to our counterparts. It indicates the effectiveness yet simpleness of our image-level alignment design that turns a naive CD model adapting cross-resolution scenarios.

% 可视化结果；比较各个数据集上的可视化效果；说明我们的方法对于跨分辨率情况适应性更好；举几个例子：目标模糊的情况下，我们的方法也能很好的检测到，同时注意与base方法比较。

\textbf{Visual comparison}. Fig. \ref{fig:comparison_sota} illustrates the visual results of the compared change detection model on the LEVIR-CD(4$\times$), SV-CD(8$\times$), and DE-CD(3.3$\times$) test sets under the fixed cross-resolution setting. We use different colors to denote TP (white), TN (black), FP (red), and FN (green). Note that results of some early pure-convolutional CD models (FC-EF, FC-Siam-Conc, and FC-Siam-Diff) are not included for a better view. We can observe that the proposed model could achieve better predictions across the three datasets. For instance, as shown in Fig. \ref{fig:comparison_sota} (b) where three new build-ups appear on the left of the region, conventional CD methods are struggling to recognize these changes of interest due to their weak textures caused by downsampling the post-event image. SUNet tends to overestimate the change areas resulting in relatively lower precision. Our method could yield relatively accurate results despite blurred regions that occurred changes. It may be due to our designed change decoder that learns implicitly the shape of changes by using dense coordinate querying an INR MLP, therefore recovering HR changes of interest even if given LR degraded inputs.

\textbf{Comparison of model efficiency.} To a fair comparison, all the models are trained and tested on a computing server equipped with a single NVIDIA RTX 3090 GPU. Table \ref{tab:model_efficiency} reports the number of model parameters (Params.), floating-point operations per second (FLOPs), and GPU training time of each method. The input to the model has a size of $256 \times 256 \times 3$. The reported time corresponds to the duration required to complete one epoch of training on the LEVIR-CD dataset using a batch size of 8. The results show that the proposed method outperforms the recent DMINet and ICIFNet with smaller model parameters and less computational cost. From Table \ref{tab:model_efficiency} and Table \ref{tab:comparison_sota}, we can observe that the proposed method achieves significant accuracy improvement compared to our baseline while utilizing a modest increase in model parameters and maintaining acceptable computational consumption.

\begin{table}
    \centering
    \caption{Comparison of model efficiency. We report the number of model parameters (Params.), floating-point operations per second (FLOPs), and training time for one epoch on the LEVIR-CD training set. The input image to the model has a size of $256\times 256 \times 3$. The batch size is set to 8.}
    % \resizebox{0.3\textwidth}{!}{
    \begin{tabular}{c|ccc}
    \toprule
    Model & Params.(M) & FLOPs (G) & Training Time (s) \\
    \midrule
        FC-EF \cite{Daudt2018} & 1.35 &  7.14 &  52.33 \\
        FC-Siam-conc \cite{Daudt2018} & 1.55 &  10.64 &  64.99 \\
        FC-Siam-diff \cite{Daudt2018} & 1.35 &  9.44  &  63.21 \\
        STANet \cite{Chen2020e} & 16.93 & 26.32 &   270.74 \\
        IFNet \cite{Zhang2020b} & 50.71 &  164.70 & 372.66 \\ 
        SNUNet \cite{Fang2021} & 12.03 & 109.76 & 305.82 \\
        BIT \cite{Chen2022} & 3.55 & 17.42 &  140.46 \\
        ICIFNet \cite{Feng2022} & 25.83 & 50.50 & 555.37 \\
        DMINet \cite{Feng2023} &  6.76 & 28.38 & 172.29 \\
        \midrule
        SUNet \cite{Shao2021} & 15.56  & 79.78 & 321.31 \\
        SRCDNet \cite{Liu2022g} & 12.77 & 30.98 & 81.86 \\
        
    \midrule
        Base &  11.97 & 11.42 & 53.18 \\
        Ours & 13.06 & 17.5 & 103.81 \\
    \bottomrule
    \end{tabular}
    % }
    \label{tab:model_efficiency}
\end{table}

\subsection{Handling Continuous Resolution Difference Ratios}

To further verify the model adaptation ability for continuous cross-resolution conditions, we feed samples of varying resolution difference ratios ($r_{d}$) across bitemporal images into the CD model that are trained on a fixed difference ratio setting. For a fair comparison, we apply image-space alignment by interpolating the LR image to an HR reconstruction before feeding it to each CD model. 

Let $r_{d0}$ be the original resolution difference ratio of the training samples. $r_{d0}$ equals 4, 8, and 3.3 for LEVIR-CD, SV-CD, and DE-CD datasets, respectively.
Based on the resolution difference ratio in the validation/testing phase compared to that during training, we primarily have two settings: in-distribution and out-of-distribution settings. For simplicity, we denote values between 1 to $r_{d0}$ as in-distribution ratios, and those larger than $r_{d0}$ as out-of-distribution ratios.  
Given one HR bitemporal sample from LEVIR-CD and SV-CD datasets, we downsample the post-event HR image with different scales to obtain samples with varying ratios. For the real-world DE-CD dataset, because of the lack of real HR pre-event images, we downsample the post-event HR image for in-distribution conditions and further downsample the pre-event LR image for out-of-distribution conditions.

Table \ref{tab:comparison_cr_levir}, Table \ref{tab:comparison_cr_sv}, and Table \ref{tab:comparison_cr_de} report the cross-resolution performance of different CD models on the LEIVR-CD, SV-CD, and DE-CD test sets, respectively. Quantitive results show our proposed method not only consistently outperforms other methods in terms of F1/IoU scores across the three datasets in the in-distribution settings, but also exhibits significant advantages in the out-of-distribution settings.

We can observe that most of the methods achieve optimal results under a certain in-distribution ratio, while in the out-of-distribution setting, as the resolution difference ratio increases, the performance decreases. It is not surprising that the optimal ratio of most methods is less than the difference ratio of the training data. It is because these models train the Siamese encoder to adapt to both HR and LR data, which may result in a compromise of an in-between resolution. It is worth noting that the proposed method exhibits nearly consistent performance for each in-distribution setting, ranging from 1$\times$ to 8$\times$, on the SV-CD dataset, while our base model on the 1$\times$ setting is much inferior to (i.e., 1.3 points of the F1 score drops) that on the 8$\times$ setting. It may be attributed to our design of the scale-invariant learning framework as well as the change decoder which implicitly represents the detailed shape of land covers of interest. We can also observe that the proposed method achieves larger performance boosts compared to other methods in the case of out-of-distribution compared to in-distribution settings. For example, in the LEVIR-CD test set, our method significantly outperforms the counterpart (e.g., DMINet) by 16 points in terms of F1 score in the out-of-distribution (8$\times$) setting, compared to 1.3 points in the in-distribution (4$\times$) setting.
Moreover, we can observe that some early approaches, e.g., FC-EF, FC-Siam-Conc, FC-Siam-Diff, and SUNet somehow exhibit relatively insufficient yet stable performance across different resolution differences. Some recent advanced CD methods such as DMINet and ICIFNet deliver promising performance in scenarios with small resolution differences but their performance declines significantly in cases of the large resolution difference settings (e.g., over 20 percent drops in terms of F1 score on the LEVIR-CD dataset of 8$\times$ setting). It may be because these methods tend to overfit the known patterns and struggle to adapt to unseen ones. Overall, the proposed method demonstrates a balanced performance, consistently outperforming others across all cross-resolution settings.

\begin{figure*}
\begin{minipage}[t]{0.33\linewidth}
\centering
\includegraphics[width=\textwidth]{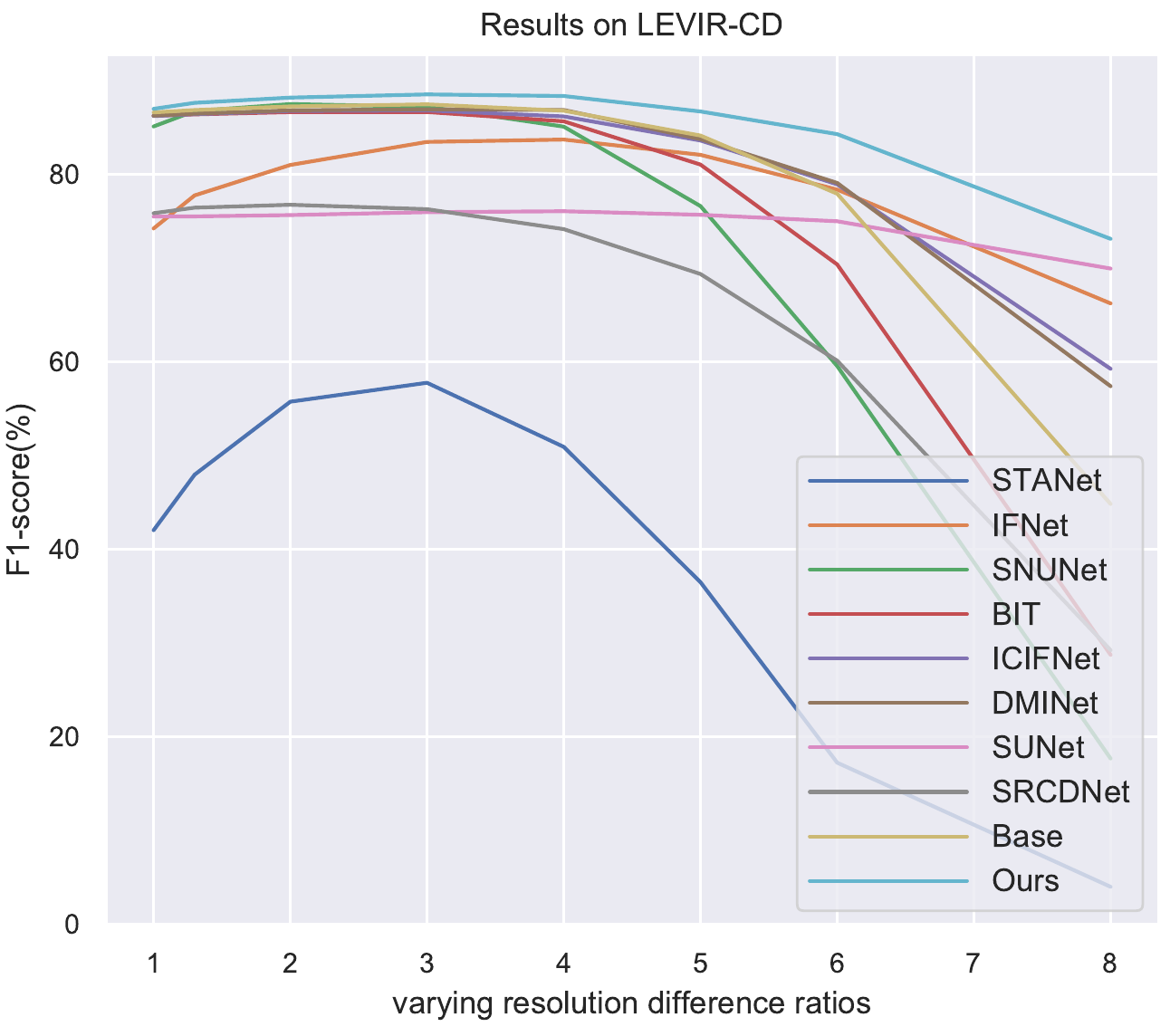}
  \centerline{(a) LEVIR-CD}
\end{minipage}
\begin{minipage}[t]{0.33\linewidth}
\centering
\includegraphics[width=\textwidth]{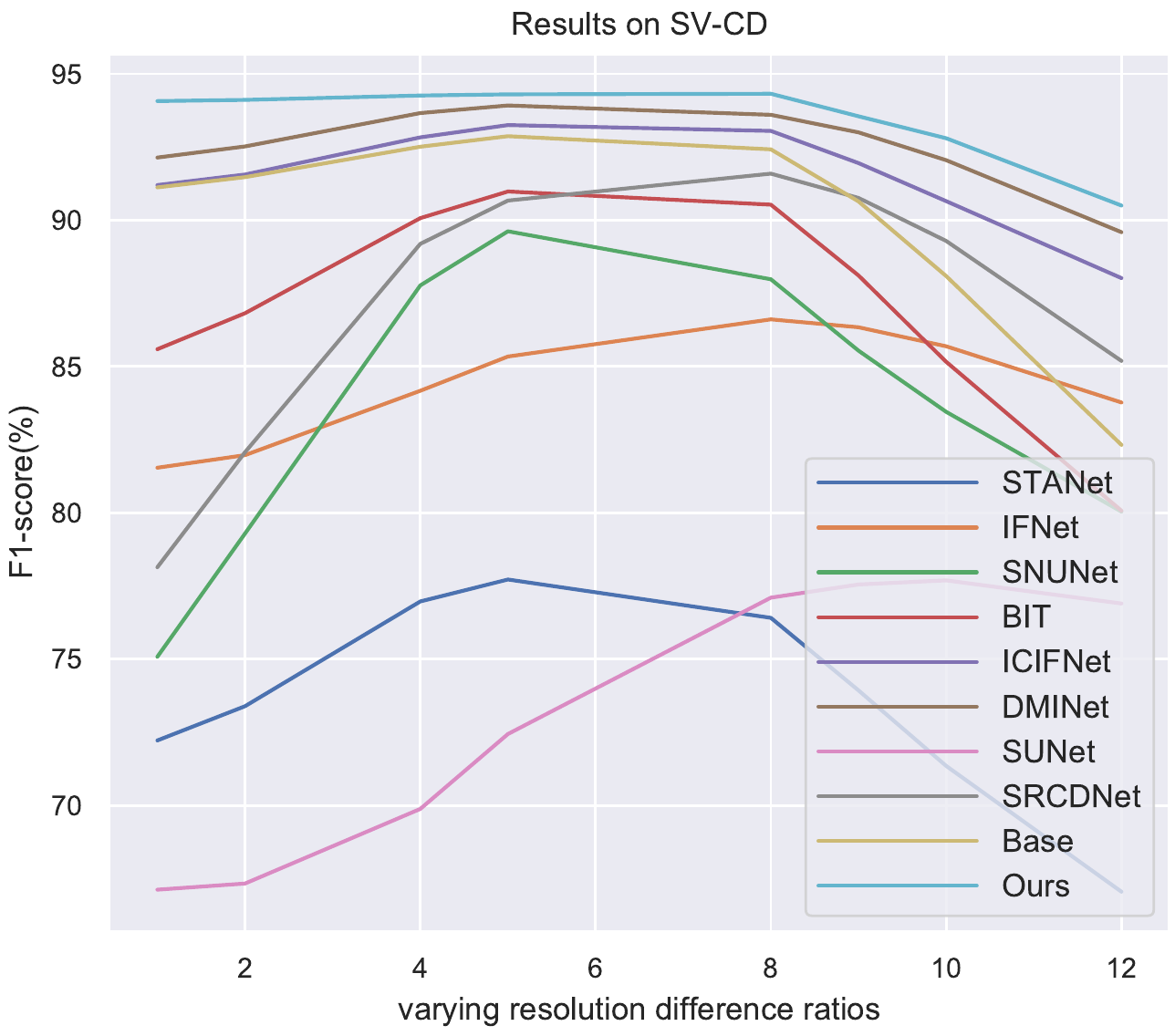}
  \centerline{(b) SV-CD}
\end{minipage}
\begin{minipage}[t]{0.33\linewidth}
\centering
\includegraphics[width=\textwidth]{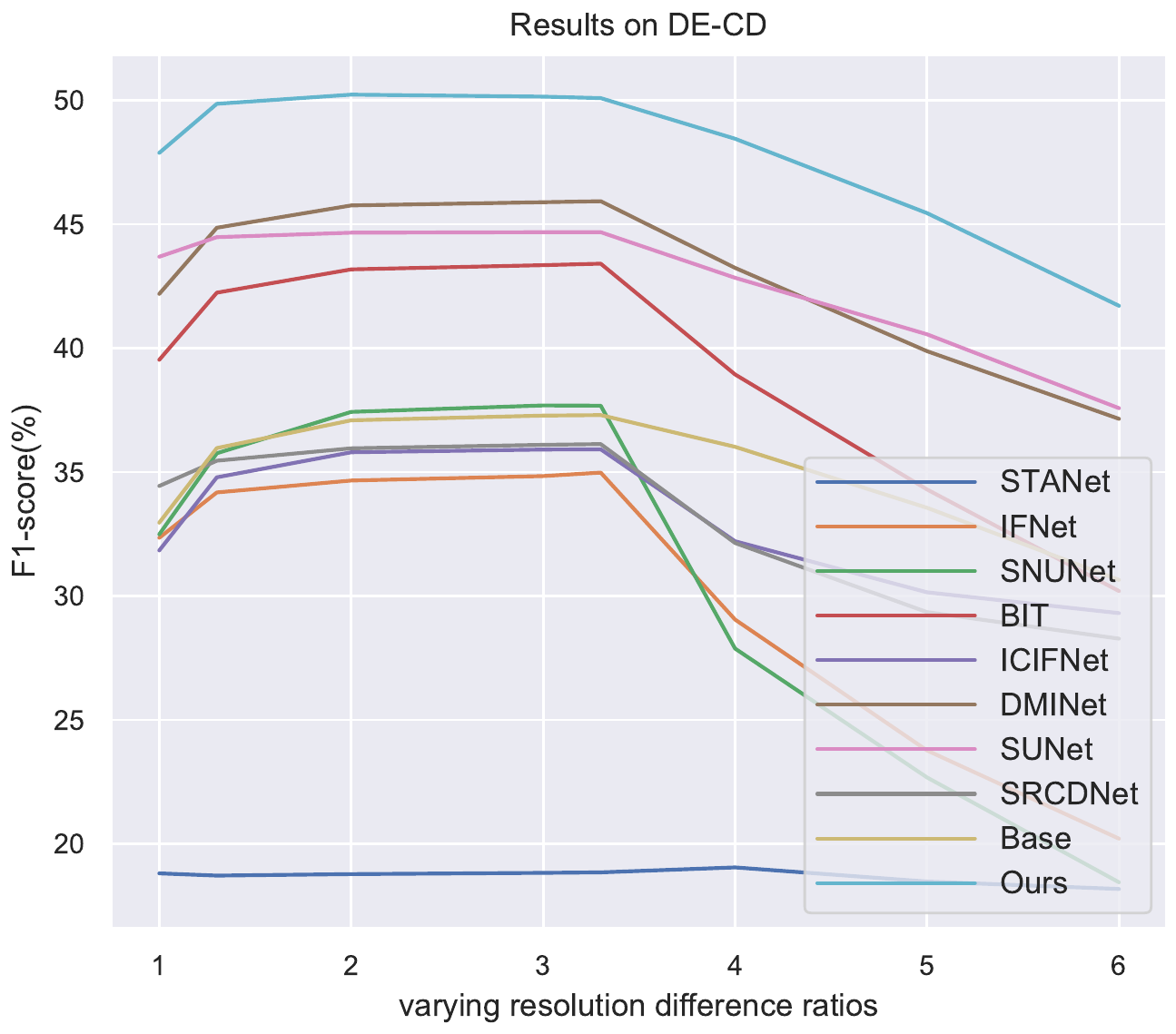}
  \centerline{(c) DE-CD}
\end{minipage}
\caption{F1-score comparison using varying bitemporal resolution difference ratios on the LEVIR-CD, SV-CD, and DE-CD test sets, respectively. The F1-score is reported.}
\label{fig:comparison_curve}
\end{figure*}

To better illustrate the cross-resolution adaptability of our method, we display the F1-score curve of different models under varying resolution difference ratios on the LEVIR-CD, SV-CD, and DE-CD test sets in Fig. \ref{fig:comparison_curve}. We can observe that our method substantially shows more stability and better accuracy than other methods.

\begin{figure*}
        \centering
        \includegraphics[width=1\textwidth]{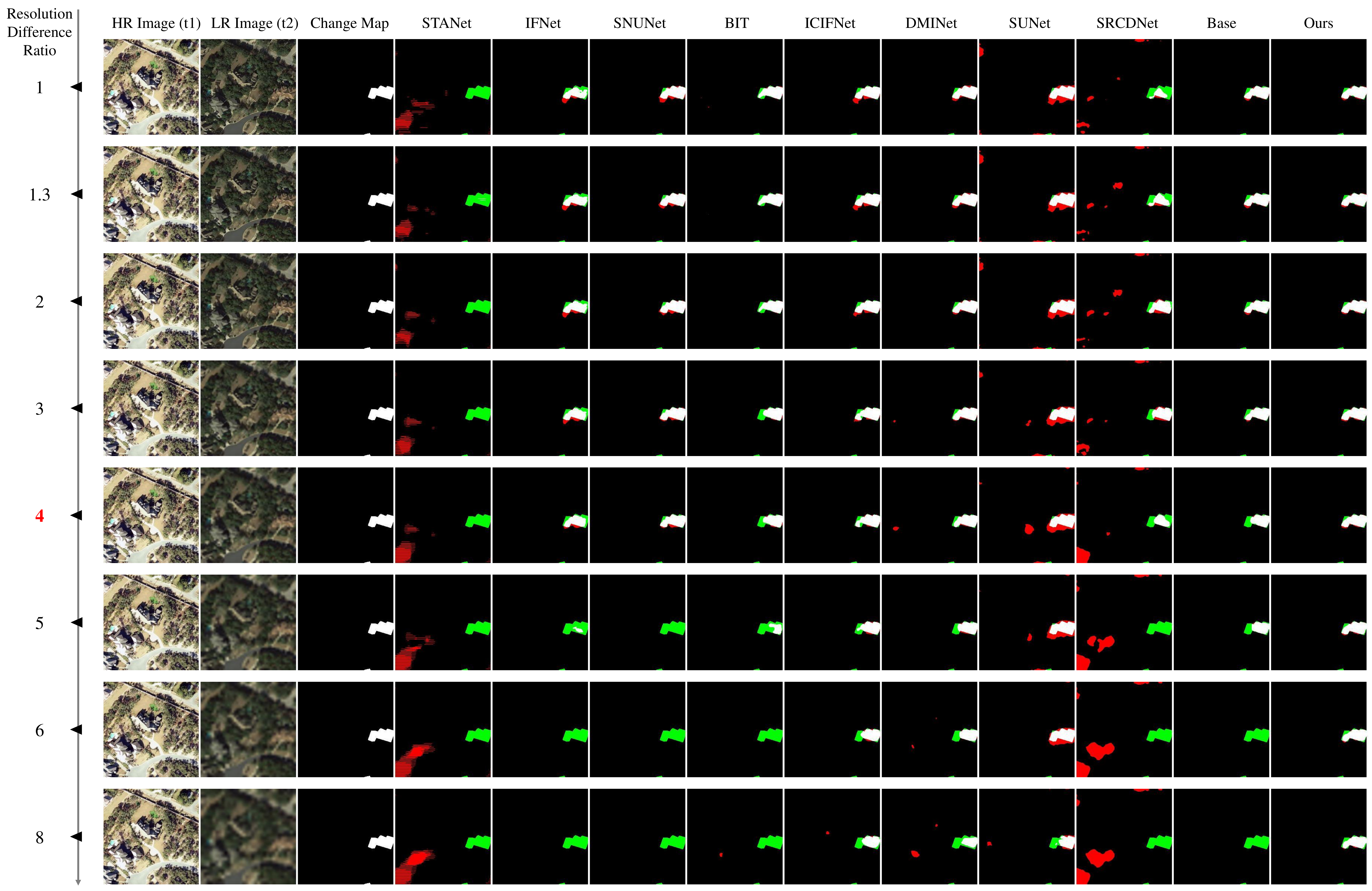}
        \caption{Visual comparison of different methods on the LEVIR-CD test set with varying bitemporal resolution difference ratios. We synthesize LR images by downsampling with different scales the post-event image. For a better view, we use white for true positive, black for true negative, red for false positive, and green for false negative. 
        }
        \label{fig:comparison_continous_levir}
\end{figure*}

\begin{figure*}
        \centering
        \includegraphics[width=1\textwidth]{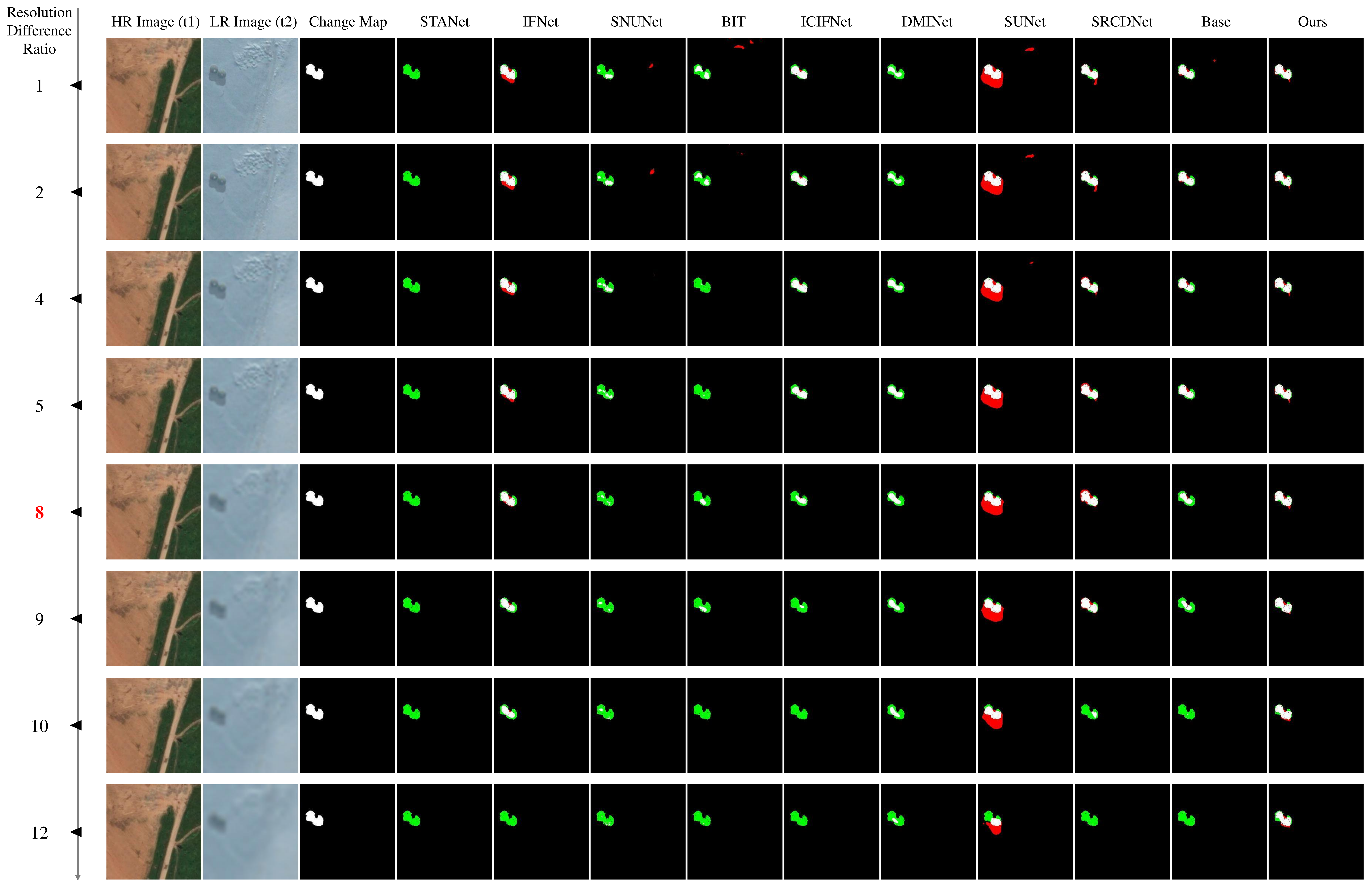}
        \caption{Visual comparison of different methods on the SV-CD test set with varying bitemporal resolution difference ratios. We synthesize LR images by downsampling with different scales the post-event image. For a better view, we use white for true positive, black for true negative, red for false positive, and green for false negative.
        }
        \label{fig:comparison_continous_sv}
\end{figure*}

\begin{figure*}
        \centering
        \includegraphics[width=1\textwidth]{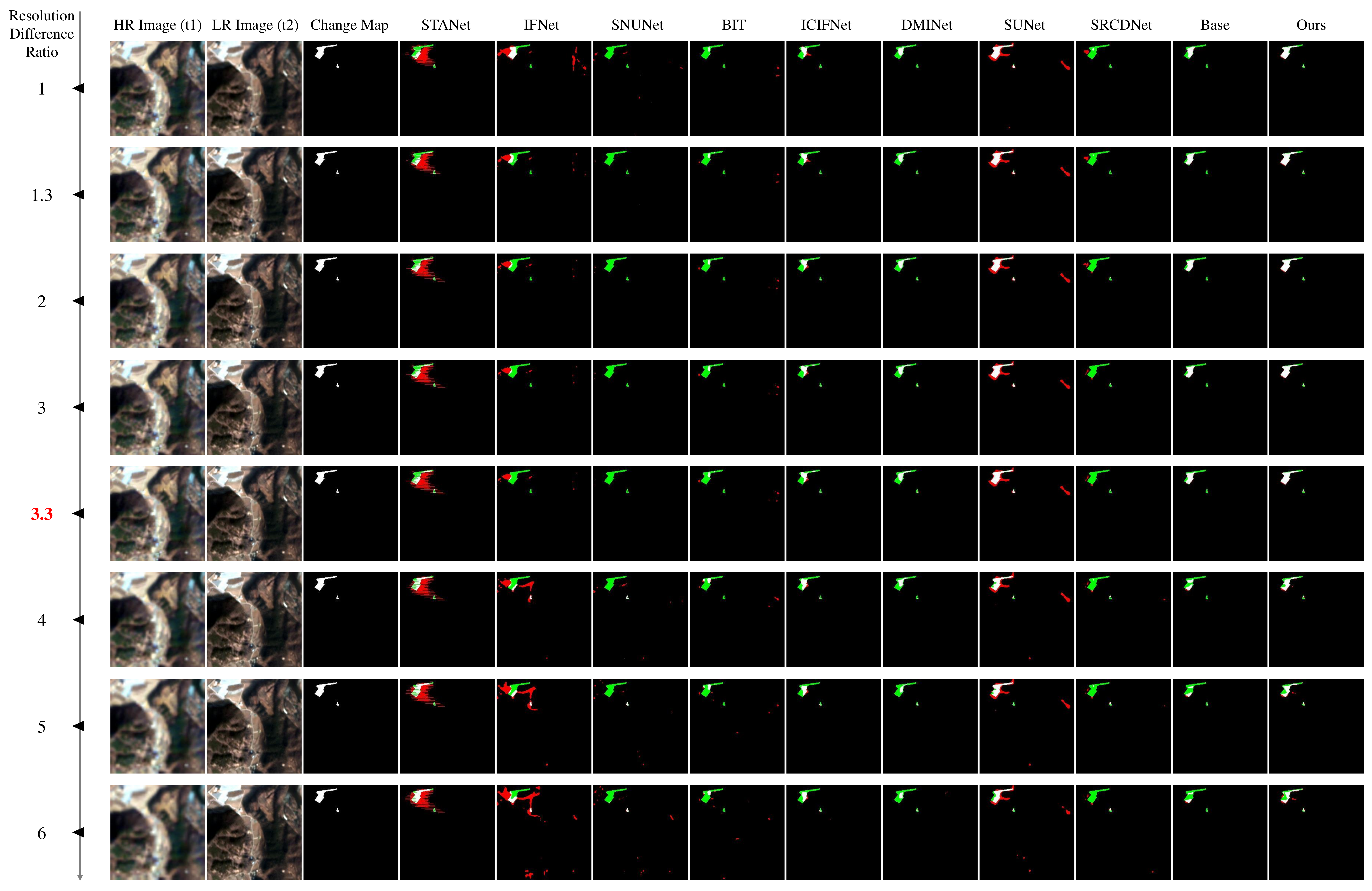}
        \caption{Visual comparison of different methods on the DE-CD test set with varying bitemporal resolution difference ratios. Due to lacking real HR pre-event images, we synthesize relatively HR images by downsampling the post-event image, for in-distribution conditions. The real LR pre-event image is further downsampled for out-of-distribution testing. For a better view, we use white for true positive, black for true negative, red for false positive, and green for false negative.
        }
        \label{fig:comparison_continous_de}
\end{figure*}

Fig. \ref{fig:comparison_continous_levir}, Fig. \ref{fig:comparison_continous_sv}, and Fig. \ref{fig:comparison_continous_de} also illustrate the visual results of compared models on these datasets with varying bitemporal resolution difference ratios. The visual comparison also verifies the cross-resolution adaptability of the proposed method. For instance, Fig. \ref{fig:comparison_continous_sv} shows some newly built ground facilities on the left side of the region. Our method can obtain consistent accurate predictions across varying difference ratios while most other compared methods fail to recognize the change of interest under the out-of-distribution ratios (e.g., 12$\times$).

\begin{table*}
    \centering
    \caption{Cross-resolution comparison on the LEVIR-CD test set with varying bitemporal resolution difference ratios. We synthesize LR images by downsampling post-event images. The best results for each cross-resolution setting are marked in \textbf{bold}. All the scores are described as percentages (\%). These models are trained on the samples from the LEVIR-CD(4$\times$) training set.}
    \resizebox{1\textwidth}{!}{
    \begin{tabular}{c|ccccc|ccc}
  \toprule
    \multicolumn{1}{c}{} &
    \multicolumn{5}{|c|}{In-distribution testing (F1 / IoU)} &
    \multicolumn{3}{|c}{Out-of-distribution testing (F1 / IoU)}\\
    \multicolumn{1}{c}{} &
    \multicolumn{1}{|c}{\textbf{1$\times$}}  &
    \multicolumn{1}{c}{\textbf{1.3$\times$}}  &
    \multicolumn{1}{c}{\textbf{2$\times$}} &
    \multicolumn{1}{c}{\textbf{3$\times$}} &
    \multicolumn{1}{c}{\textbf{4$\times$}} &
    \multicolumn{1}{|c}{\textbf{5$\times$}} &
    \multicolumn{1}{c}{\textbf{6$\times$}} &
    \multicolumn{1}{c}{\textbf{8$\times$}} \\
    \midrule
    FC-EF \cite{Daudt2018} 
    & 73.67 / 58.31 & 74.49 / 59.35 & 75.11 / 60.14 & 75.49 / 60.62 & 75.31 / 60.40 & 74.41 / 59.24 & 72.52 / 56.88 & 66.83 / 50.18 \\
    FC-Siam-Conc \cite{Daudt2018} 
    & 78.63 / 64.78 & 79.33 / 65.74 & 79.32 / 65.73 & 78.34 / 64.39 & 76.40 / 61.81 & 73.55 / 58.16 & 69.91 / 53.73 & 61.57 / 44.48 \\
    FC-Siam-Diff \cite{Daudt2018} 
    & 75.39 / 60.51 & 76.29 / 61.66 & 76.06 / 61.36 & 74.17 / 58.94 & 70.83 / 54.83 & 66.12 / 49.39 & 60.35 / 43.22 & 46.57 / 30.35  \\
    STANet \cite{Chen2020e} 
    & 42.03 / 26.61 & 47.96 / 31.54 & 55.75 / 38.64 & 57.77 / 40.62 & 50.93 / 34.16 & 36.49 / 22.32 & 17.22 / 9.42 & 3.97 / 2.02 \\
    IFNet \cite{Zhang2020b}  
    & 74.24 / 59.03 & 77.77 / 63.62 & 81.02 / 68.10 & 83.48 / 71.65 & 83.73 / 72.01 & 82.10 / 69.64 & 78.39 / 64.47 & 66.24 / 49.52 \\
    SNUNet \cite{Fang2021} 
    & 85.13 / 74.11 & 86.77 / 76.62 & 87.52 / 77.81 & 87.24 / 77.36 & 85.11 / 74.09 & 76.63 / 62.12 & 59.56 / 42.41 & 17.66 / 9.68 \\
    BIT \cite{Chen2022}  
    & 86.28 / 75.86 & 86.42 / 76.09 & 86.66 / 76.47 & 86.67 / 76.48 & 85.68 / 74.94 & 81.06 / 68.16 & 70.40 / 54.33 & 28.73 / 16.78 \\
    ICIFNet \cite{Feng2022}  
    &  86.24 / 75.80 & 86.44 / 76.11 & 86.78 / 76.65 & 86.84 / 76.75 & 86.20 / 75.75 & 83.63 / 71.87 & 78.95 / 65.22 & 59.26 / 42.10 \\
    DMINet \cite{Feng2023}  
    & 86.28 / 75.87 & 86.49 / 76.20 & 86.85 / 76.75 & 86.96 / 76.93 & 86.89 / 76.82 & 83.78 / 72.08 & 79.10 / 65.43 & 57.40 / 40.26 \\ 
    \midrule
    SUNet \cite{Shao2021}  
    &  75.51 / 60.65 & 75.53 / 60.69 & 75.67 / 60.86 & 75.98 / 61.27 & 76.08 / 61.40 & 75.70 / 60.90 & 75.01 / 60.02 & 69.96 / 53.80 \\
    SRCDNet \cite{Liu2022g}  
    & 75.87 / 61.12 & 76.46 / 61.89 & 76.77 / 62.30 & 76.30 / 61.68 & 74.17 / 58.94 & 69.38 / 53.12 & 60.13 / 42.99 & 29.23 / 17.12 \\
    \midrule
    Base
    & 86.63 / 76.42 & 86.88 / 76.81 & 87.27 / 77.41 & 87.49 / 77.76 & 86.81 / 76.70 & 84.16 / 72.65 & 77.95 / 63.87 & 44.83 / 28.89 \\
    Ours 
    & \textbf{87.01} / \textbf{77.01} & \textbf{87.65} / \textbf{78.02} & \textbf{88.21} / \textbf{78.90} & \textbf{88.55} / \textbf{79.44} & \textbf{88.38} / \textbf{79.18} & \textbf{86.73} / \textbf{76.57} & \textbf{84.31} / \textbf{72.87} & \textbf{73.13} / \textbf{57.64} \\
   \bottomrule
    \end{tabular}
    }
    \label{tab:comparison_cr_levir}
\end{table*}

\begin{table*}
    \centering
    \caption{Cross-resolution comparison on the SV-CD test set with varying bitemporal resolution difference ratios. We synthesize LR images by downsampling post-event images. The best results for each cross-resolution setting are marked in \textbf{bold}. All the scores are described as percentages (\%). These models are trained on the samples from the SV-CD(8$\times$) training set.}
    \resizebox{1\textwidth}{!}{
    \begin{tabular}{c|ccccc|ccc}
  \toprule
    \multicolumn{1}{c}{} &
    \multicolumn{5}{|c|}{In-distribution testing (F1 / IoU)} &
    \multicolumn{3}{|c}{Out-of-distribution testing (F1 / IoU)}\\
    \multicolumn{1}{c}{} &
    \multicolumn{1}{|c}{\textbf{1$\times$}}  &
    \multicolumn{1}{c}{\textbf{2$\times$}}  &
    \multicolumn{1}{c}{\textbf{4$\times$}} &
    \multicolumn{1}{c}{\textbf{5$\times$}} &
    \multicolumn{1}{c}{\textbf{8$\times$}} &
    \multicolumn{1}{|c}{\textbf{9$\times$}} &
    \multicolumn{1}{c}{\textbf{10$\times$}} &
    \multicolumn{1}{c}{\textbf{12$\times$}} \\
    \midrule
    FC-EF \cite{Daudt2018} 
    & 55.88 / 38.77 & 55.88 / 38.77 & 55.96 / 38.85 & 55.99 / 38.88 & 56.06 / 38.95 & 56.06 / 38.94 & 56.06 / 38.95 & 56.03 / 38.92  \\
    FC-Siam-Conc \cite{Daudt2018} 
    & 64.31 / 47.39 & 64.52 / 47.62 & 63.97 / 47.03 & 63.35 / 46.36 & 59.99 / 42.85 & 58.39 / 41.23 & 56.84 / 39.70 & 53.88 / 36.87  \\
    FC-Siam-Diff \cite{Daudt2018} 
    & 67.35 / 50.77 & 67.41 / 50.84 & 67.09 / 50.48 & 66.86 / 50.22 & 64.97 / 48.12 & 63.54 / 46.56 & 61.84 / 44.76 & 58.51 / 41.35 \\
    STANet \cite{Chen2020e} 
    & 72.22 / 56.52 & 73.39 / 57.97 & 76.97 / 62.56 & 77.72 / 63.55 & 76.41 / 61.82 & 73.93 / 58.64 & 71.35 / 55.47 & 67.05 / 50.43 \\
    IFNet \cite{Zhang2020b}  
    & 81.54 / 68.83 & 81.97 / 69.45 & 84.17 / 72.66 & 85.34 / 74.42 & 86.61 / 76.38 & 86.34 / 75.97 & 85.69 / 74.97 & 83.77 / 72.07 \\
    SNUNet \cite{Fang2021} 
    & 75.08 / 60.10 & 79.29 / 65.69 & 87.77 / 78.21 & 89.62 / 81.20 & 87.98 / 78.55 & 85.54 / 74.74 & 83.45 / 71.60 & 80.04 / 66.73 \\
    BIT \cite{Chen2022}  
    & 85.59 / 74.81 & 86.82 / 76.70 & 90.07 / 81.94 & 90.98 / 83.46 & 90.53 / 82.69 & 88.11 / 78.75 & 85.16 / 74.15 & 80.07 / 66.77 \\
    ICIFNet \cite{Feng2022}  
    &  91.20 / 83.83 & 91.56 / 84.44 & 92.83 / 86.63 & 93.25 / 87.35 & 93.05 / 87.00 & 91.95 / 85.10 & 90.65 / 82.90 & 88.02 / 78.60 \\
    DMINet \cite{Feng2023}  
    & 92.14 / 85.42 & 92.52 / 86.09 & 93.66 / 88.07 & 93.92 / 88.54 & 93.60 / 87.96 & 93.00 / 86.91 & 92.05 / 85.27 & 89.59 / 81.14  \\ 
    \midrule
    SUNet \cite{Shao2021}  
    &  67.12 / 50.51 & 67.33 / 50.76 & 69.88 / 53.70 & 72.44 / 56.80 & 77.10 / 62.73 & 77.55 / 63.33 & 77.69 / 63.52 & 76.90 / 62.46  \\
    SRCDNet \cite{Liu2022g}  
    & 78.14 / 64.13 & 82.07 / 69.59 & 89.19 / 80.49 & 90.67 / 82.93 & 91.59 / 84.49 & 90.76 / 83.08 & 89.29 / 80.66 & 85.19 / 74.19  \\
    \midrule
    Base
    & 91.12 / 83.69 & 91.47 / 84.28 & 92.51 / 86.06 & 92.87 / 86.68 & 92.42 / 85.90 & 90.64 / 82.88 & 88.09 / 78.71 & 82.32 / 69.96 \\
    Ours 
    & \textbf{94.07} / \textbf{88.80} & \textbf{94.11} / \textbf{88.88} & \textbf{94.26} / \textbf{89.14} & \textbf{94.30} / \textbf{89.22} & \textbf{94.32} / \textbf{89.24} & \textbf{93.55} / \textbf{87.87} & \textbf{92.80} / \textbf{86.57} & \textbf{90.50} / \textbf{82.65} \\
   \bottomrule
    \end{tabular}
    }
    \label{tab:comparison_cr_sv}
\end{table*}

\begin{table*}
    \centering
    \caption{Cross-resolution comparison on the DE-CD test set with varying bitemporal resolution difference ratios. For in-distribution testing, we synthesize relatively HR images compared to real pre-event LR images by downsampling post-event images. For out-of-distribution testing, we further downsample pre-event images to synthesize LR images. The best results for each cross-resolution setting are marked in \textbf{bold}. All the scores are described as percentages (\%). These models are trained on the samples from the DE-CD(3.3$\times$) training set.}
    \resizebox{1\textwidth}{!}{
    \begin{tabular}{c|ccccc|ccc}
  \toprule
    \multicolumn{1}{c}{} &
    \multicolumn{5}{|c|}{In-distribution testing (F1 / IoU)} &
    \multicolumn{3}{|c}{Out-of-distribution testing (F1 / IoU)}\\
    \multicolumn{1}{c}{} &
    \multicolumn{1}{|c}{\textbf{1$\times$}}  &
    \multicolumn{1}{c}{\textbf{1.3$\times$}}  &
    \multicolumn{1}{c}{\textbf{2$\times$}} &
    \multicolumn{1}{c}{\textbf{3$\times$}} &
    \multicolumn{1}{c}{\textbf{3.3$\times$}} &
    \multicolumn{1}{|c}{\textbf{4$\times$}} &
    \multicolumn{1}{c}{\textbf{5$\times$}} &
    \multicolumn{1}{c}{\textbf{6$\times$}} \\
    \midrule
    STANet \cite{Chen2020e} 
    & 18.81 / 10.38 & 18.72 / 10.33 & 18.78 / 10.37 & 18.82 / 10.39 & 18.85 / 10.41 & 19.05 / 10.53 & 18.48 / 10.18 & 18.18 / 10.00\\
    IFNet \cite{Zhang2020b}  
    & 32.35 / 19.30 & 34.18 / 20.62 & 34.66 / 20.96 & 34.84 / 21.09 & 34.98 / 21.20 & 29.05 / 17.00 & 23.77 / 13.49 & 20.21 / 11.24  \\
    SNUNet \cite{Fang2021} 
    & 32.49 / 19.40 & 35.76 / 21.77 & 37.43 / 23.02 & 37.69 / 23.22 & 37.68 / 23.21 & 27.88 / 16.20 & 22.68 / 12.79 & 18.45 / 10.16  \\
    BIT \cite{Chen2022}  
    & 39.53 / 24.63 & 42.24 / 26.78 & 43.18 / 27.53 & 43.35 / 27.68 & 43.41 / 27.72 & 38.94 / 24.18 & 34.30 / 20.70 & 30.20 / 17.78 \\
    ICIFNet \cite{Feng2022}  
    &  31.84 / 18.93 & 34.79 / 21.06 & 35.80 / 21.80 & 35.91 / 21.89 & 35.92 / 21.89 & 32.21 / 19.20 & 30.15 / 17.75 & 29.31 / 17.17 \\
    DMINet \cite{Feng2023}  
    & 31.60 / 18.77 & 30.89 / 18.27 & 30.26 / 17.83 & 30.09 / 17.71 & 30.02 / 17.66 & 29.31 / 17.17 & 27.30 / 15.81 & 23.74 / 13.47  \\ 
    \midrule
    SUNet \cite{Shao2021}  
    &  43.69 / 27.95 & 44.48 / 28.60 & 44.66 / 28.75 & 44.68 / 28.77 & 44.68 / 28.77 & 42.84 / 27.25 & 40.56 / 25.44 & 37.58 / 23.14  \\
    SRCDNet \cite{Liu2022g}  
    & 34.44 / 20.80 & 35.46 / 21.55 & 35.96 / 21.92 & 36.10 / 22.02 & 36.13 / 22.05 & 32.14 / 19.14 & 29.35 / 17.20 & 28.28 / 16.47  \\
    \midrule
    Base
    & 32.96 / 19.73 & 35.97 / 21.93 & 37.09 / 22.76 & 37.28 / 22.91 & 37.30 / 22.93 & 36.02 / 21.97 & 33.56 / 20.16 & 30.65 / 18.10  \\
    Ours 
    & \textbf{47.88} / \textbf{31.48}  & \textbf{49.86} / \textbf{33.21} & \textbf{50.23} / \textbf{33.54} & \textbf{50.15} / \textbf{33.54} & \textbf{50.09} / \textbf{33.42} & \textbf{48.45} / \textbf{31.97} & \textbf{45.45} / \textbf{29.41} & \textbf{41.71} / \textbf{26.35} \\
   \bottomrule
    \end{tabular}
    }
    \label{tab:comparison_cr_de}
\end{table*}

\begin{table*}
    \centering
    \caption{Cross-resolution comparison on the LEVIR-CD test set. These models are trained on the samples from the original (1$\times$) LEVIR-CD training set and are tested with samples with varying bitemporal resolution difference ratios. We synthesize LR images by downsampling post-event images. The F1-score and IoU are reported.}
    \resizebox{1\textwidth}{!}{
    \begin{tabular}{c|cccccccc}
  \toprule
    \multicolumn{1}{c|}{} &
    \multicolumn{1}{c}{\textbf{1$\times$}}  &
    \multicolumn{1}{c}{\textbf{1.3$\times$}}  &
    \multicolumn{1}{c}{\textbf{2$\times$}} &
    \multicolumn{1}{c}{\textbf{3$\times$}} &
    \multicolumn{1}{c}{\textbf{4$\times$}} &
    \multicolumn{1}{c}{\textbf{5$\times$}} &
    \multicolumn{1}{c}{\textbf{6$\times$}} &
    \multicolumn{1}{c}{\textbf{8$\times$}} \\
    \midrule
    FC-EF \cite{Daudt2018} 
    & 75.79 / 61.01 & 75.78 / 61.01 & 75.43 / 60.56 & 74.46 / 59.32 & 72.79 / 57.21 & 70.48 / 54.42 & 67.41 / 50.84 & 59.74 / 42.59 \\
    FC-Siam-Conc \cite{Daudt2018} 
    & 82.28 / 69.90 & 82.12 / 69.66 & 81.52 / 68.80 & 79.93 / 66.57 & 77.60 / 63.39 & 74.31 / 59.13 & 70.08 / 53.95 & 58.51 / 41.35 \\
    FC-Siam-Diff \cite{Daudt2018} 
    & 79.17 / 65.52 & 79.20 / 65.56 & 78.30 / 64.33 & 76.00 / 61.30 & 72.77 / 57.19 & 67.70 / 51.18 & 61.42 / 44.32 & 45.36 / 29.33  \\
    STANet \cite{Chen2020e} 
    & 87.27 / 77.41 & 86.70 / 76.53 & 85.14 / 74.12 & 72.79 / 57.22 & 40.43 / 25.34 & 16.45 / 8.96 & 9.10 / 4.77 & 7.00 / 3.62\\
    IFNet \cite{Zhang2020b}  
    & 88.11 / 78.75 & 88.06 / 78.67 & 87.42 / 77.65 & 85.31 / 74.39 & 80.95 / 68.00 & 73.35 / 57.91 & 59.29 / 42.13 & 21.01 / 11.74 \\
    SNUNet \cite{Fang2021} 
    & 89.37 / 80.78 & 89.25 / 80.58 & 88.28 / 79.01 & 77.61 / 63.41 & 44.43 / 28.56 & 15.77 / 8.56 & 9.87 / 5.19 & 8.40 / 4.38\\
    BIT \cite{Chen2022}  
    & 88.54 / 79.44 & 88.57 / 79.48 & 88.23 / 78.94 & 86.03 / 75.49 & 78.66 / 64.83 & 61.00 / 43.88 & 33.69 / 20.26 & 7.84 / 4.08  \\
    ICIFNet \cite{Feng2022}  
    &  88.20 / 78.89 & 88.18 / 78.86 & 87.92 / 78.44 & 86.48 / 76.19 & 82.01 / 69.50 & 72.62 / 57.01 & 54.95 / 37.88 & 15.92 / 8.65 \\
    DMINet \cite{Feng2023}  
    & 89.56 / 81.09 & 89.45 / 80.91 & 89.01 / 80.20 & 87.29 / 77.45 & 82.64 / 70.42 & 70.60 / 54.55 & 48.92 / 32.38 & 11.01 / 5.83  \\ 
    \midrule
    SUNet \cite{Shao2021}  
    &  78.32 / 64.37 & 78.21 / 64.22 & 77.94 / 63.85 & 77.41 / 63.14 & 76.69 / 62.19 & 75.42 / 60.53 & 73.40 / 57.97 & 61.83 / 44.75 \\
    SRCDNet \cite{Liu2022g}  
    & 76.66 / 62.15 & 76.38 / 61.78 & 75.91 / 61.17 & 73.31 / 57.87 & 64.78 / 47.91 & 51.14 / 34.35 & 30.15 / 17.75 & 7.25 / 3.76 \\
    \midrule
    Base
    & 88.63 / 79.59 & 88.60 / 79.53 & 88.23 / 78.93 & 86.29 / 75.88 & 79.25 / 65.63 & 57.95 / 40.79 & 22.08 / 12.41 & 0.64 / 0.32 \\
    Ours 
    & \textbf{89.70} / \textbf{81.33} & \textbf{89.67} / \textbf{81.27} & \textbf{89.24} / \textbf{80.58} & \textbf{88.14} / \textbf{78.80} & \textbf{85.79} / \textbf{75.11} & \textbf{81.65} / \textbf{68.99} & \textbf{75.11} / \textbf{60.14} & \textbf{65.17} / \textbf{48.33} \\
   \bottomrule
    \end{tabular}
    }
    \label{tab:comparison_cr_levir0}
\end{table*}

Apart from the setting of cross-resolution training/testing, i.e., the model is trained on samples with fixed resolution difference ratios and then validated on samples with different cross-resolution conditions, we also perform the setting of original-resolution training and cross-resolution testing, i.e., the model is trained on equal-resolution samples from the original CD training set and then validated on those with varying cross-resolution conditions.

Table \ref{tab:comparison_cr_levir0} reports the cross-resolution performance of different models on the LEVIR-CD dataset set. Each compared model is trained on the HR training samples with equal bitemporal resolution from the original LEVIR-CD dataset.
In the training phase, we perform random downsampled reconstruction on the pre-event image by a ratio from Uniform distribution $r\sim U[1, 8]$. 
Similarly, we downsample the post-event HR image using different scales to obtain cross-resolution samples in the testing phase. 
Quantitative results show that the proposed method consistently outperforms other methods in terms of F1/IoU scores on testing samples with different cross-resolution ratios. We can observe from the results that most methods achieve the best results when the ratio is equal to 1, while the performance decreases when the ratio increases. For instance, DMINet exhibits comparable performance to our method when the ratio equals 1, but when the ratio increases to 8, its performance is dramatically dropped by nearly 90 percent, while our method could maintain acceptable performance. The results further indicate the cross-resolution adaptability of the proposed method.

\subsection{Ablation Studies}
\label{ssec:ablation}

We perform ablation experiments on the three critical components of the proposed methods, i.e., Random Resolution Synthesis (RRS), Implicit Neural Decoder (IND), and Bitemporal Local Interaction (BLI). We start from the baseline (Base) and incrementally supplement the above three components to evaluate their respective gains to the CD performance. 

Table \ref{tab:abalation_sili} reports the ablation results of our method on the LEVIR-CD(4$\times$), SV-CD(8$\times$), and DE-CD(3.3$\times$) test sets. The F1-score of each model is listed for comparison. Quantitative results show that the three components of SILI bring consistent performance improvements across different datasets.

\textbf{Ablation on RRS}. As shown in Table \ref{tab:abalation_sili}, compared to baseline, random resolution synthesis brings in significant improvements across the three datasets. 
It is not surprising because such a design can be viewed as a data augmentation approach, that synthesizes degraded reconstructions with various intrinsic resolutions. 
For the cross-resolution CD task, our data-level design synthesizing images with intermediate resolutions between low- and high-resolution inputs may benefit model learning by providing a progression of resolutions to reduce the resolution gap across bitemporal images.

\textbf{Ablation on IND}. We can observe from Table \ref{tab:abalation_sili} that our proposed INR-based change decoder could further consistently improve the baseline on the three datasets, especially on the DE(3.3$\times$) dataset with relatively lower spatial resolution and with smaller pixel numbers per change area. It indicates the effectiveness of our IND in yielding the HR change mask from multi-level features, especially for recovering small objects of change. We further make a comparison to several conventional multi-level feature fusion approaches, including FPN \cite{Lin2017a} and UNet \cite{Ronneberger2015}. 
Those structures perform incremental aggregation from coarse to fine for the multi-level features (level $1$ to level $4$) from each temporal image. The concatenated bitemporal HR semantic features are then fed into three-layer convolutions for change classification, similar to our baseline. Quanuantive results in Table \ref{tab:abalation_inr} suggest the effectiveness of our IND for the cross-resolution CD task, compared with counterparts. Note that each model in Table \ref{tab:abalation_inr} is trained with RRS for a fair comparison.

\textbf{Ablation on BLI}. Table \ref{tab:abalation_sili} demonstrates that our bitemporal local interaction produces consistent performance gains across the three datasets. To further demonstrate the effectiveness of BLI, we also compare the commonly used global self-attention. For a fair comparison, we only replace the local-window self-attention in BLI with the global self-attention. Table \ref{tab:abalation_bli} shows the comparison results on the three datasets. Note that here we only add bitemporal interaction on the features of level 1 from the encoder.  Quantitative results show that our method consistently outperforms the self-attention counterpart, suggesting that local bitemporal interactions are more effective for the cross-resolution CD task. It indicates that modeling spatial-temporal correlations in the local regions between cross-resolution bitemporal images may be sufficient to align their semantic features.

\begin{table}
    \centering
    \caption{Ablation study of our SILI on three CD datasets. Ablations are performed on the Random Resolution Synthesis (RRS), Implicit Neural Decoder (IND), and Bitemporal Local Interaction (BLI). The F1-score is reported.}
    % \resizebox{0.3\textwidth}{!}{
    \begin{tabular}{ccccccc}
    \toprule
    Model & RRS & IND & BLI & LEVIR(4$\times$) & SV(8$\times$) & DE(3.3$\times$) \\
    \midrule
    Base & $\times$ & $\times$ & $\times$  & 86.81 & 92.42 & 37.30  \\ 
    \midrule 
    SILI & $\checkmark$ & $\times$ & $\times$ & 87.48 & 93.49 & 40.73 \\
    SILI & $\checkmark$ & $\checkmark$ & $\times$  & 88.04 & 94.16 & 48.86 \\
    SILI & $\checkmark$ & $\checkmark$ & $\checkmark$ & \textbf{88.38} & \textbf{94.32} & \textbf{50.17} \\
    \bottomrule
    \end{tabular}
    % }
    \label{tab:abalation_sili}
\end{table}

\begin{table}
    \centering
    \caption{Effect of our INR-based change decoder. We replace INR with several off-the-shell multi-level feature fusion approaches for comparison. The F1/IoU score of each model on three CD datasets is reported. }
    % \resizebox{0.3\textwidth}{!}{
    \begin{tabular}{cccc}
    \toprule
    % Model & deocder & LEVIR(4$\times$) & SV(8$\times$) & DE(3.3$\times$) \\
    \multicolumn{1}{c}{} &
    \multicolumn{1}{c}{\textbf{LEVIR}(4$\times$)}  &  \multicolumn{1}{c}{\textbf{SV}(8$\times$)} & \multicolumn{1}{c}{\textbf{DE}(3.3$\times$)} \\
    Decoder & F1 / IoU  & F1 / IoU & F1 / IoU  \\
    \midrule
    FPN & 85.96 / 75.37 & 93.31 / 87.45 & 39.48 / 24.60 \\ 
    UNet & 87.78 / 78.22 & 93.41 / 87.63 & 43.99 / 28.20\\
    MLP & 87.48 / 77.74 & 93.49 / 87.77 & 40.73 / 25.58 \\
    \midrule
    INR & \textbf{88.04} / \textbf{78.64} & \textbf{94.16} / \textbf{89.87} & \textbf{48.86} / \textbf{32.33}\\
    \bottomrule
    \end{tabular}
    % }
    \label{tab:abalation_inr}
\end{table}

\begin{table}
    \centering
    \caption{Effect of random bitemporal region swap on three CD datasets. We also perform ablations on the size of the swapped region. The F1/IoU scores of each model are reported. Note that a crop size of 0 denotes not performing region swap. A crop size of 256 means to swap the bitemporal image, i.e., the whole region of the image. We use our Base model as the baseline.}
    % \resizebox{0.3\textwidth}{!}{
    \begin{tabular}{cccc}
    \toprule
    % Edge (/ds) & LEVIR(4$\times$) & SV(8$\times$) & DE(3.3$\times$) \\
    \multicolumn{1}{c}{} &
    \multicolumn{1}{c}{\textbf{LEVIR}(4$\times$)}  &  \multicolumn{1}{c}{\textbf{SV}(8$\times$)} & \multicolumn{1}{c}{\textbf{DE}(3.3$\times$)} \\
    Crop size & F1 / IoU  & F1 / IoU & F1 / IoU  \\
    \midrule
    $0 $ & 86.81 / 76.70  & 92.42 / 85.90 & 37.30 / 22.93 \\ 
    \midrule 
    $64 $ & 87.13 / 77.19 & 92.81 / 86.58 & 37.96 / 23.43 \\
    $128 $  & \textbf{87.15} / \textbf{77.23} & \textbf{92.94} / \textbf{86.81} & \textbf{38.15} / \textbf{23.57} \\
    $192 $ & 87.04 / 77.06 & 92.86 / 86.66 & 37.49 / 23.43 \\
    \midrule
    $256 $ & 86.89 / 76.82 & 92.69 / 86.37 & 36.94 / 22.65 \\
    \bottomrule
    \end{tabular}
    % }
    \label{tab:abalation_swap}
\end{table}

\begin{table}
    \centering
    \caption{Effect of the local-window attention in the bitemporal feature interaction. We replace local attention with non-local self-attention for comparison. Note that we only apply interaction on the bitemporal features of level 1 from the encoder. The F1/IoU score of each model on three CD datasets is reported. }
    % \resizebox{0.3\textwidth}{!}{
    \begin{tabular}{cccc}
    \toprule
    % interaction & LEVIR(4$\times$) & SV(8$\times$) & DE(3.3$\times$) \\
    \multicolumn{1}{c}{} &
    \multicolumn{1}{c}{\textbf{LEVIR}(4$\times$)}  &  \multicolumn{1}{c}{\textbf{SV}(8$\times$)} & \multicolumn{1}{c}{\textbf{DE}(3.3$\times$)} \\
    interaction & F1 / IoU  & F1 / IoU & F1 / IoU  \\
    \midrule
    $\times$ & 88.04 / 78.64 & 94.16 / 89.87 & 48.86 / 32.33  \\ 
    \midrule 
    non-local & 88.12 / 78.76 & 93.87 / 88.44 & 48.90 / 32.36 \\
    local & \textbf{88.24} / \textbf{78.96}& \textbf{94.23} / \textbf{89.09} & \textbf{49.33} / \textbf{32.74}\\
    \bottomrule
    \end{tabular}
    % }
    \label{tab:abalation_bli}
\end{table}

\begin{table}
    \centering
    \caption{Effect of the resolution of dense coordinate queries in the change decoder on three CD datasets. We also perform ablations on whether to introduce edge features. The FLOPs and F1/IoU scores of each model are reported. Note that ds denotes the downsampling rate of the coordinate query map related to the original HR image.}
    % \resizebox{0.3\textwidth}{!}{
    \begin{tabular}{ccccc}
    \toprule
    % Edge (/ds) & LEVIR(4$\times$) & SV(8$\times$) & DE(3.3$\times$) \\
    \multicolumn{1}{c}{} & \multicolumn{1}{c}{} &
    \multicolumn{1}{c}{\textbf{LEVIR}(4$\times$)}  &  \multicolumn{1}{c}{\textbf{SV}(8$\times$)} & \multicolumn{1}{c}{\textbf{DE}(3.3$\times$)} \\
    Edge (/ds) & FLOPs (G) & F1 / IoU  & F1 / IoU & F1 / IoU  \\
    \midrule
    $\times (/4)$ & 11.52 & 87.38 / 77.56  & 93.65 / 88.06 & 45.72 / 29.63 \\ 
    learn (/4) & 11.54 & 87.60 / 77.93  & 93.81 / 88.33 & 45.25 / 29.24 \\ 
    \midrule 
    $\checkmark (/4)$ & 11.54 & 87.67 / 78.04 & 93.94 / 88.57 & 48.40 / 31.93 \\
    $\checkmark (/2)$ & 17.20 & 88.04 / 78.64 & \textbf{94.16} / \textbf{89.87} & 48.86 / 32.33 \\
    $\checkmark (/1)$ & 39.84 & \textbf{88.05} / \textbf{78.65} & 93.97 / 88.62 & \textbf{49.36}	/ \textbf{32.77} \\
    \bottomrule
    \end{tabular}
    % }
    \label{tab:abalation_edge}
\end{table}

\begin{table}
    \centering
    \caption{Effect of introducing bitemporal interaction at different stages (from level 1 to level 4) of the encoder. The F1-score of each model on three CD datasets is reported.}
    % \resizebox{0.3\textwidth}{!}{
    \begin{tabular}{ccccccc}
    \toprule
    1 & 2 & 3 & 4 & LEVIR(4$\times$) & SV(8$\times$) & DE(3.3$\times$) \\
    \midrule
    $\times$ & $\times$ & $\times$ & $\times$  & 88.04 & 94.16 & 48.86  \\ 
    \midrule 
    $\checkmark$ & $\times$ & $\times$ & $\times$ & 88.24 & 94.23 & 49.33 \\
    $\checkmark$ & $\checkmark$ & $\times$ & $\times$  & 88.34 & 94.18 & 49.41 \\
    $\checkmark$ & $\checkmark$ & $\checkmark$ & $\times$ & \textbf{88.38} & 94.32 & \textbf{50.10} \\
    $\checkmark$ & $\checkmark$ & $\checkmark$ & $\checkmark$ & 88.34 & \textbf{94.35} & 49.75 \\
    \bottomrule
    \end{tabular}
    % }
    \label{tab:effect_bil_level}
\end{table}

\subsection{Parametric Analysis}
\label{subsec:parameter}

\textbf{Effect of Random Bitemporal Region Swap.} We propose to swap a random region between bitemporal images with different intrinsic spatial resolutions as a form of patch-level data augmentation to benefit the learning of scale-invariant features. The size of the swapped region, i.e., crop size, is an important hyperparameter. To explore the impact of crop size on CD performance, we perform ablation on different crop sizes for bitemporal region swapping. Our Base model is used as the baseline. Table \ref{tab:abalation_swap} reports the F1/IoU scores of compared models with different crop sizes. Note that the crop size of 0 denotes not applying the bitemporal region swap. The crop size of 256 means to swap the entire image in the temporal dimension, which is equivalent to not using region swap because bitemporal images do not interact with each other at the image level. Quantitative results show that the model with random region swap significantly outperforms the baseline. It indicates the effectiveness of the proposed random bitemporal region swap. This approach can be regarded as a form of patch-level data augmentation through the interaction of bitemporal information. Notably, the optimal results are attained with a crop size of 128, with a slight performance decrease observed as the crop size increases to 192. This reduction in performance with larger crop sizes is attributed to the increased likelihood of foreground land covers appearing at the swap area's edges, introducing truncated and incomplete land cover instances that can impede the model's learning process. Therefore, we set the crop size to 128.

\textbf{Effect of the resolution of coordinate query map}. 
Our INR-based change decoder uses dense coordinate queries alongside corresponding multi-level features to obtain the HR change mask. The spatial resolution of the coordinate query map is an important hyperparameter. Let $ds$ be the downsampling factor of the coordinate query map relative to the original HR image. Note that we directly bilinearly interpolate the relatively LR change prediction from the decoder to match the size of the HR ground truth when applying LR coordinate queries.
Table \ref{tab:abalation_edge} reports the floating-point operations per second (FLOPs), and F1/IoU scores of compared models with different $ds$. Note that here we use our SILI model without BIL for experiments. From the last three rows of the table, we can observe that when the resolution of queries increases, model performance on the three datasets improves overall, yet with higher computational complexity. For a trade-off between accuracy and efficiency, we set $ds=2$.
Additionally, we also verified the effectiveness of introducing edge clues. Quantitative results in Table \ref{tab:abalation_edge} manifest adding edge clues can consistently improve the model performance on the three datasets. To further validate the efficacy of incorporating handicraft edge features, we conduct a comparison between the models with and without these features. Note that we set up two baselines, the first baseline model (i.e., $\times$(/4)) does not receive any image edge features. The second baseline (i.e., learn (/4)) utilizes a learnable convolution layer to extract edge features from each temporal image and subsequently aggregate them to derive edge clues. For a fair comparison, the second baseline has the same amount of additional convolution parameters as our model does. Quantitative results in Table \ref{tab:abalation_edge} show that introducing additional edge features could consistently improve the CD performance in the three datasets. It indicates the effectiveness of the incorporation of handicraft edge features and learnable features, which has also been witnessed in some recent works \cite{zheng2022hfa, liu2022idan, shangguan2023contour}. It may be because the introduction of handicraft edge features could offer additional high-frequency information that may benefit network optimization.

\textbf{Which stages to introducing BLI}. We introduce BLI on bitemporal image features from a certain stage of the encoder. Here, we explore which stages to introduce bitemporal interactions. We choose our SILI model without any BLI as the baseline and incrementally add bitemporal interactions from level 1 to level 4. As shown in the table \ref{tab:effect_bil_level}, as the number of bitemporal interactions increases, the performance of the model in terms of F1 score broadly progressively improves. Concretely, BLI brings in significant performance gains across the three datasets in the early stages of the encoder, while in the last stage (level 4), introducing BLI achieves relatively limited improvement, or even degrades the performance. It may be because the feature discrepancy caused by the difference in radiation and intrinsic resolution between bitemporal images could be better aligned by BLI during the early stages.
Therefore, our SILI introduces interactions in stages of level 1/2/3.

\subsection{Feature Visualization}
Here, we provide an example to visualize multi-level features from our model to further demonstrate the effectiveness of introducing BLI. We use a popular feature visualization technique, class activation map (CAM) \cite{Wang2020f}, to show what our model learns in each stage of the encoder. CAM  is basically the channel-wise weighted sum of activation maps from a certain layer in the model. We visualize the last layer of each stage in the encoder.

\begin{figure*}
        \centering
        \includegraphics[width=1\textwidth]{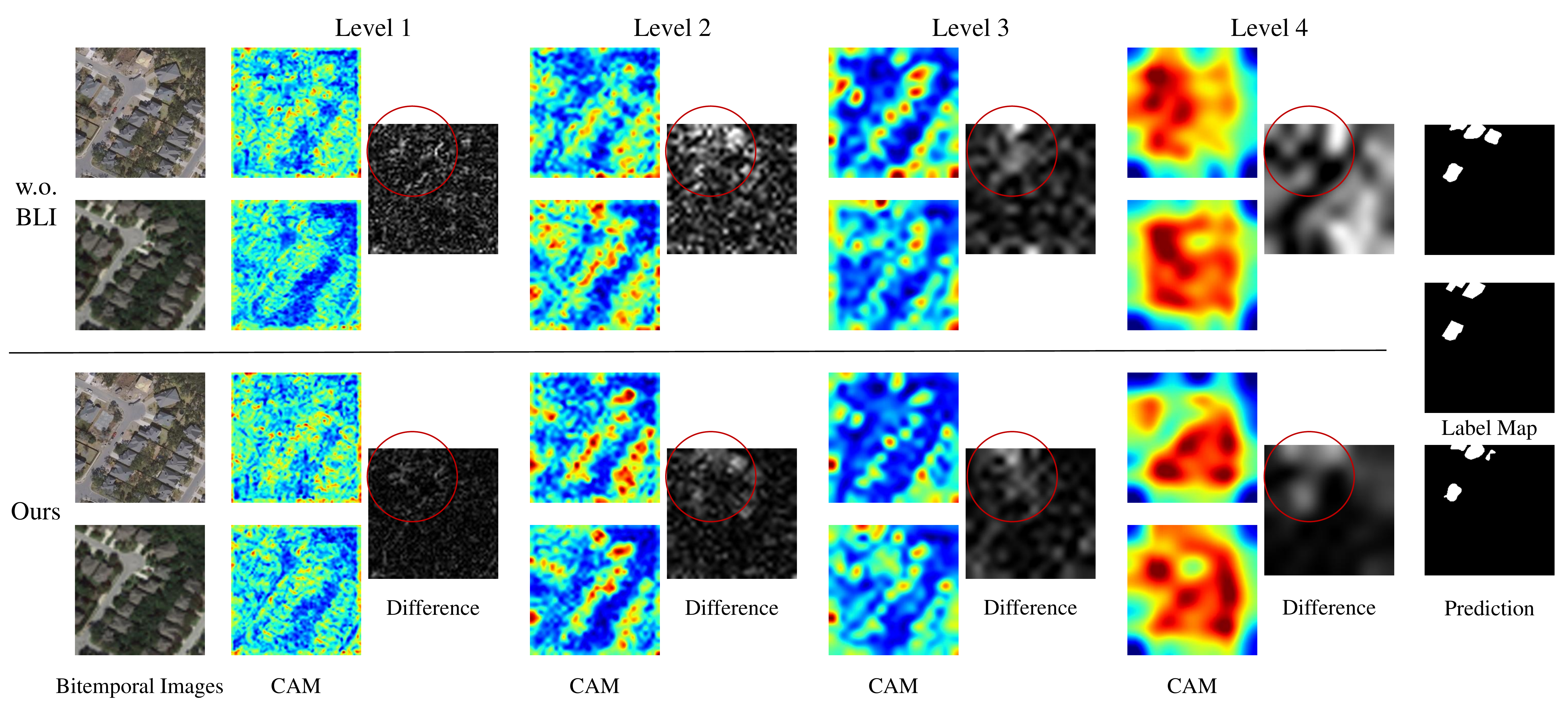}
        \caption{Feature visualization of our models with or without bitemporal interactions. We show the class activation map (CAM) for each temporal image from level 1 to level 4. Bitemporal feature difference is also displayed to better show the effectiveness of introducing BLI. The input sample is from LEVIR-CD ($4\times$) test set.
        }
        \label{fig:feature_vis}
\end{figure*}

Fig. \ref{fig:feature_vis} shows the CAM visualization of our models with or without BLI. Red denotes high values while blue denotes low values. The input sample is from LEVIR-CD ($4\times$) test set. We can observe from the CAM of each level that our model can concentrate on land covers on interest (building). Features from level 1 contain more spatial details, and those from level 4 are more semantic information but lack location precision while the intermediate levels (2/3) provide a balanced representation that well localizes semantic elements. 
We can also observe that our method with BLI has similar intensities between bitemporal features of no-change regions. We further show feature difference maps, i.e. absolute subtraction between bitemporal unnormalized CAMs. We can observe that positions with high bitemporal difference values of our model are mainly distributed within the red box, while the model without BLI may exhibit large difference values (e.g., level 2/4) outside the red box where contains no changes. It suggests the effectiveness of BLI in aligning bitemporal semantic features and yielding relatively lower feature differences in regions of no change.

\section{Conclusion}
\label{sec:conclusion}

In this paper, we propose a scale-invariant method with implicit neural networks to achieve continuous cross-resolution RS image CD. The scale-invariant embedding space is learned by enforcing our model predicting the HR change mask given synthesized bitemporal images with random downsampling and region swapping. Dense coordinate queries and corresponding multi-level features are used for change recognition by an MLP that implicitly represents the shape of changes. Bitemporal local interaction is further introduced at early levels of the encoder to align bitemporal feature intensities regardless of resolution differences. Extensive experiments on two synthesized and one real-world cross-resolution CD datasets verify the effectiveness of the proposed method. Our SILI significantly outperforms several conventional CD methods and two specifically designed cross-resolution CD methods on the three datasets in both in-distribution and out-of-distribution settings.
Our method could yield relatively consistent HR change predictions regardless of the resolution difference between bitemporal images. The empirical results suggest that our method could well handle varying bitemporal resolution difference ratios, towards real-world applications. Future works include, 1) exploring more effective scale-invariant change detection methods from the perspective of model architecture by incorporating scale-invariant network structures, rather than indirectly enhancing scale invariance through multiscale data augmentation, 2) investigating more advanced implicit neural representation techniques and their integration into the change detection task to achieve resolution-invariant change detection, 3) exploring the combination of various handcrafted features such as LBP, HOG, with deep learning models to evaluate their potential for improving CD performance.

%  ******************************appendices******************************  
% \appendices

% \section{}

%  ******************************Acknowledgment******************************   
% use section* for Acknowledgment
% \section*{Acknowledgment}


\ifCLASSOPTIONcaptionsoff
  \newpage
\fi

%  ******************************references section****************************** 
{\small
\bibliographystyle{IEEEtran}
\bibliography{references}
}

\end{document}